\documentclass{article}

\usepackage{microtype}
\usepackage{graphicx}
\usepackage{multirow}
\usepackage{amsthm}
\usepackage{amssymb}
\usepackage{mathtools}
\usepackage{amsmath}
\usepackage{booktabs}
\usepackage{subcaption}
\usepackage{float}
\usepackage{enumitem}
\usepackage{hyperref}

\usepackage[accepted]{icml2025}

\usepackage{amsmath}
\usepackage{amssymb}
\usepackage{mathtools}
\usepackage{amsthm}
\usepackage{csquotes}
\DeclareMathOperator*{\argmin}{arg\,min}

\usepackage[capitalize,noabbrev]{cleveref}

\theoremstyle{plain}

\theoremstyle{definition}

\theoremstyle{remark}

\usepackage[textsize=tiny]{todonotes}

\icmltitlerunning{Feature Shift Localization Network}

\begin{document}

\twocolumn[
\icmltitle{Feature Shift Localization Network}

\icmlsetsymbol{equal}{*}

\begin{icmlauthorlist}
\icmlauthor{M\'{i}riam Barrab\'{e}s}{equal,1,2}
\icmlauthor{Daniel Mas Montserrat}{equal,1}
\icmlauthor{Kapal Dev}{2}
\icmlauthor{Alexander G. Ioannidis}{1,3}
\end{icmlauthorlist}

\icmlaffiliation{1}{Department of Biomedical Data Science, Stanford University, Stanford, CA 94305 USA}
\icmlaffiliation{2}{Department of Computer Science, Munster Technological University, Cork T12 P928, Ireland}
\icmlaffiliation{3}{Department of Biomolecular Engineering, University of California Santa Cruz, Santa Cruz, CA 95060 USA}

\icmlcorrespondingauthor{Alexander G. Ioannidis}{ioannidis@stanford.edu}

\icmlkeywords{Feature shift localization, shift detection, shift, feature shift, distribution shift}

\vskip 0.3in
]

\printAffiliationsAndNotice{\icmlEqualContribution}

\begin{abstract}
Feature shifts between data sources are present in many applications involving healthcare, biomedical, socioeconomic, financial, survey, and multi-sensor data, among others, where unharmonized heterogeneous data sources, noisy data measurements, or inconsistent processing and standardization pipelines can lead to erroneous features. Localizing shifted features is important to address the underlying cause of the shift and correct or filter the data to avoid degrading downstream analysis. While many techniques can detect distribution shifts, localizing the features originating them is still challenging, with current solutions being either inaccurate or not scalable to large and high-dimensional datasets. In this work, we introduce the Feature Shift Localization Network (FSL-Net), a neural network that can localize feature shifts in large and high-dimensional datasets in a fast and accurate manner. The network, trained with a large number of datasets, learns to extract the statistical properties of the datasets and can localize feature shifts from previously unseen datasets and shifts without the need for re-training. The code and ready-to-use trained model are available at \url{https://github.com/AI-sandbox/FSL-Net}.
\end{abstract}

\section{Introduction}

Feature distribution shifts between data sources are common in many real-world applications using multi-dimensional data composed of a set of \enquote{corrupted} features (i.e., dimensions) with mismatching statistical qualities between sources. These feature shifts are prevalent in healthcare, biomedical, and life sciences datasets, where different samples are generated at different organizations (e.g., hospitals, labs), with differing lab technologies, hardware, and data processing producing unique structural biases. In clinical genomics, feature shifts can arise from heterogeneous data acquisition protocols, which may involve differences in genotyping arrays or phenotype curation procedures \cite{moreno2024polygenic}. Similar shifts are found in social sciences, streaming, and e-business applications, where combining tabular and structured data from multiple sources, regions, and times without proper homogenization steps can lead to mismatching and biased features due to incorrect data collection procedures, human entry errors, faulty standardization, or erroneous data processing \cite{barchard2011preventing, 7382218}. Similarly, multi-sensor applications in the manufacturing industry, medicinal monitoring, finance analysis, and defense can suffer feature shifts due to faulty sensors and measuring devices \cite{qian2022review, NEURIPS2023_b3cd64dd}. While numerous techniques enable pre-processing data sources to reduce feature shifts, proper data homogenization can be a challenging task requiring data-dependent and domain-specific techniques \cite{lim2018merged}. When data homogenization fails, unattended feature shifts can negatively impact downstream applications, leading to erroneous scientific results or biased machine learning models, which makes feature shift localization critical in many data-driven domains.

\textit{Feature shift localization} is the task of enumerating which features of multi-dimensional datasets are originating the distribution shift between two or more data sources. The localization step is necessary to identify and correct the error source, whether by data removal or correction in tabular data-based applications or through physical intervention in multi-sensory scenario applications. Extensive literature on anomaly detection and distribution shift detection \cite{yu2018request, pan2020adversarial} has led to numerous tools for automated shift detection that are common in data-centric AI (DCAI) and machine learning systems (MLSys) technologies, providing functionalities for data quality control, homogenization, and monitoring \cite{ginart2022mldemon, piano2022detecting, zha2023data, subasri2023diagnosing}. While many methods focus on asserting whether two datasets follow the same distribution, most do not localize the exact features causing the shift, and recent promising shift localization techniques still fail to scale to large and high-dimensional datasets common in many areas \cite{kulinski2020feature, NEURIPS2023_b3cd64dd}.

In this work, we introduce a novel neural network, the \textit{Feature Shift Localization Network} (FSL-Net), which can localize shifts with high accuracy while scaling to high-dimensional and large datasets. The network extracts statistical descriptors from two datasets and then processes them to localize the features originating the distribution shift. Namely, FSL-Net has two subnetworks, a \textit{Statistical Descriptor Network} that compresses datasets into statistical functionals that summarize their underlying distribution, and a \textit{Prediction Network} that combines the statistical descriptors between datasets to predict the probability of being \textit{corrupted} for each feature. FSL-Net makes use of convolutional and pooling layers to achieve \textit{invariance} to sample order and approximate \textit{equivariance} to feature order. The network is trained end-to-end using multiple datasets with different types of simulated feature shifts and is evaluated on previously unseen datasets and shift types, showing that it generalizes well out-of-the-box to a large variety of data and shifts without the need for re-training. FSL-Net surpasses previous feature shift localization methods in localization accuracy and speed.

Our contributions include:
(1) we propose a novel neural network architecture that provides \textit{invariance} to sample order and \textit{equivariance} to feature order while scaling to large and high-dimensional datasets; 
(2) we design a training approach leading to a network that generalizes to unseen data and shifts without the need for re-training;
(3) we provide an in-depth experimental evaluation with multiple manipulation types, datasets, and network configurations.

\section{Related Work}\label{related_work}
\textbf{Distribution Shift Detection.} The detection of distribution shifts consists of predicting if $p \neq q$, where $p$ and $q$ are the reference and query distributions, respectively. Numerous techniques exist for detecting distribution shifts in univariate distributions \cite{gama2014survey, lu2018learning, pan2020adversarial}, and there is a growing focus on multivariate data \cite{rabanser2019failing}, which can exhibit various types of shifts such as marginal, concept, covariate, or label shifts \cite{lu2016concept, losing2016knn, liu2020diverse}. Recent shift detection techniques include \cite{yu2018request}, which makes use of hypothesis testing for concept drift detection, and \cite{rabanser2019failing}, which applies two-sample multivariate hypothesis testing via Maximum-Mean Discrepancy (MMD) \cite{gretton2012kernel}, univariate hypothesis tests with marginal Kolmogorov-Smirnov (KS) tests, and dimensionality reduction techniques.

\textbf{Feature Shift Localization.} While distribution shift detection techniques focus on detecting whether a shift exists between distributions, feature shift localization methods aim to predict which features are causing the shift. A notable contribution in this area is the work by \cite{kulinski2020feature}, which introduces a conditional test capable of accurately identifying shifted features with model-free and model-based approaches: K-Nearest Neighbors with KS statistic (KNN-KS), multivariate Gaussian with KS (MB-KS), multivariate Gaussian and Fisher-divergence test statistics (MB-SM), and deep density neural models with Fisher-divergence test (Deep-SM). DataFix \cite{NEURIPS2023_b3cd64dd} is a more recent method that improves localization accuracy by iteratively training a random forest to distinguish between reference and query distributions, removing the features with the highest impurity-based importance scores until divergence is minimized, and applying a knee-detection algorithm to determine the optimal stopping point. Although DataFix performs well in many cases, it struggles with detecting challenging feature shifts and scales poorly with high-dimensional and large datasets. Its repeated use of random forest training results in significant computational overhead, limiting its applicability to real-world applications involving massive datasets. In this paper, we adopt the same evaluation benchmark as DataFix and introduce FSL-Net to overcome these limitations.

\textbf{Feature Selection.} Feature selection methods localize the most relevant features for classification or regression, providing interpretability and removing redundancy. Wrapper \cite{maldonado2009wrapper, mustaqeem2017wrapper}, filtering \cite{nasir2020pearson, hopf2021filter}, and embedded methods \cite{tran2016pso, huang2018feature} are among the most common techniques. Wrapper methods select features of interest by training ML models and adding or removing features through a search process. Filtering methods include Mutual Information (MI) \cite{battiti1994using}, ANOVA-F test \cite{elssied2014novel}, Chi-square test \cite{bahassine2020feature}, Minimum Redundancy Maximum Relevance (MRMR) \cite{ding2005minimum, li2018feature}, and Fast-Conditional Mutual Information Maximization (FAST-CMIM) \cite{fleuret2004fast}, among others. Such methods extract statistical information from the data to rank the importance of each feature. Embedded methods rank features using built-in scores from ML models, such as logistic regression weights \cite{cheng2006logistic} or the mean decrease in impurity (Gini index) in random forests \cite{sylvester2018applications}, selecting those with the highest scores.

\textbf{Optimal Transport.} Optimal transport (OT) theory compares probability distributions by computing the minimal cost required to transform one into another, inducing a meaningful distance that reflects both global structure and the geometry of the underlying space. Its formulation as a linear programming problem \cite{kantorovitch1958translocation} connected OT to the broader field of optimization \cite{quanrud2018approximating}. Its relevance has since expanded across fields, including computer vision \cite{izquierdo2024optimal}, economics \cite{galichon2018optimal}, logistics \cite{nadal2015optimal}, and statistical inference \cite{goldfeld2024statistical}. Recent advances in scalable numerical solvers, such as entropic regularization and Sinkhorn iterations, have enabled OT to scale to high-dimensional settings and find applications in data science \cite{peyre2019computational, montesuma2024recent}, generative modeling \cite{sanjabi2018convergence}, and domain adaptation \cite{courty2016optimal}. However, while OT techniques can characterize divergences, they do not provide a direct methodology to localize divergent features, which is the main focus of this paper, and would require modifications in order to be applied for the feature shift localization task.

\textbf{Data-centric AI.} Data-centric AI (DCAI) is the paradigm that encapsulates tools and techniques aimed at improving data quality and quantity to build robust, accurate, and efficient AI systems. Unlike model-centric approaches, which prioritize refining models while working with a fixed dataset, DCAI emphasizes improving datasets through systematic and iterative processes. Key aspects include expanding datasets through data collection \cite{ghosh2023data}, annotation \cite{boecking2020interactive, caporali2023weakly}, augmentation \cite{montserrat2017training, geleta2023deep}, and integration \cite{xiaojuan2023data}, as well as refining data through cleaning \cite{krishnan2019automatic, costanzo2023data, 10795134} and feature engineering \cite{sinaci2023data, buckley2023feature}. The increasing complexity and scale of datasets have made automated methods indispensable for data refinement. A growing focus within DCAI is the localization of feature shifts \cite{zha2023data, NEURIPS2023_b3cd64dd}.

\textbf{Deep Sets, Equivariant, and Invariant Networks.} Invariance and equivariance properties are important in many applications and have proved successful in modeling physics and chemical systems, with numerous neural networks developed to have such properties \cite{benton2020learning, batzner20223, ruhe2024clifford}. Graph neural networks and attention-based networks have been shown to provide similar properties \cite{lim2022equivariant}. Deep Sets \cite{zaheer2017deep} introduced an architecture for modeling set-structured data, ensuring invariance on the order of samples. Our proposed FSL-Net adopts similar design choices as Deep Sets to obtain sample-order invariance and approximate feature-order equivariance.

Section \ref{app:benchmarking_methods} provides a more detailed description of the benchmarking methods evaluated in this paper, including DataFix, MB-SM, MB-KS, KNN-KS, and Deep-SM, as well as MI, SelectKBest, MRMR, and Fast-CMIM.

\section{Feature Shift Localization Network}

\begin{figure*}[t]
    \centering
    \includegraphics[width=1.0\textwidth]{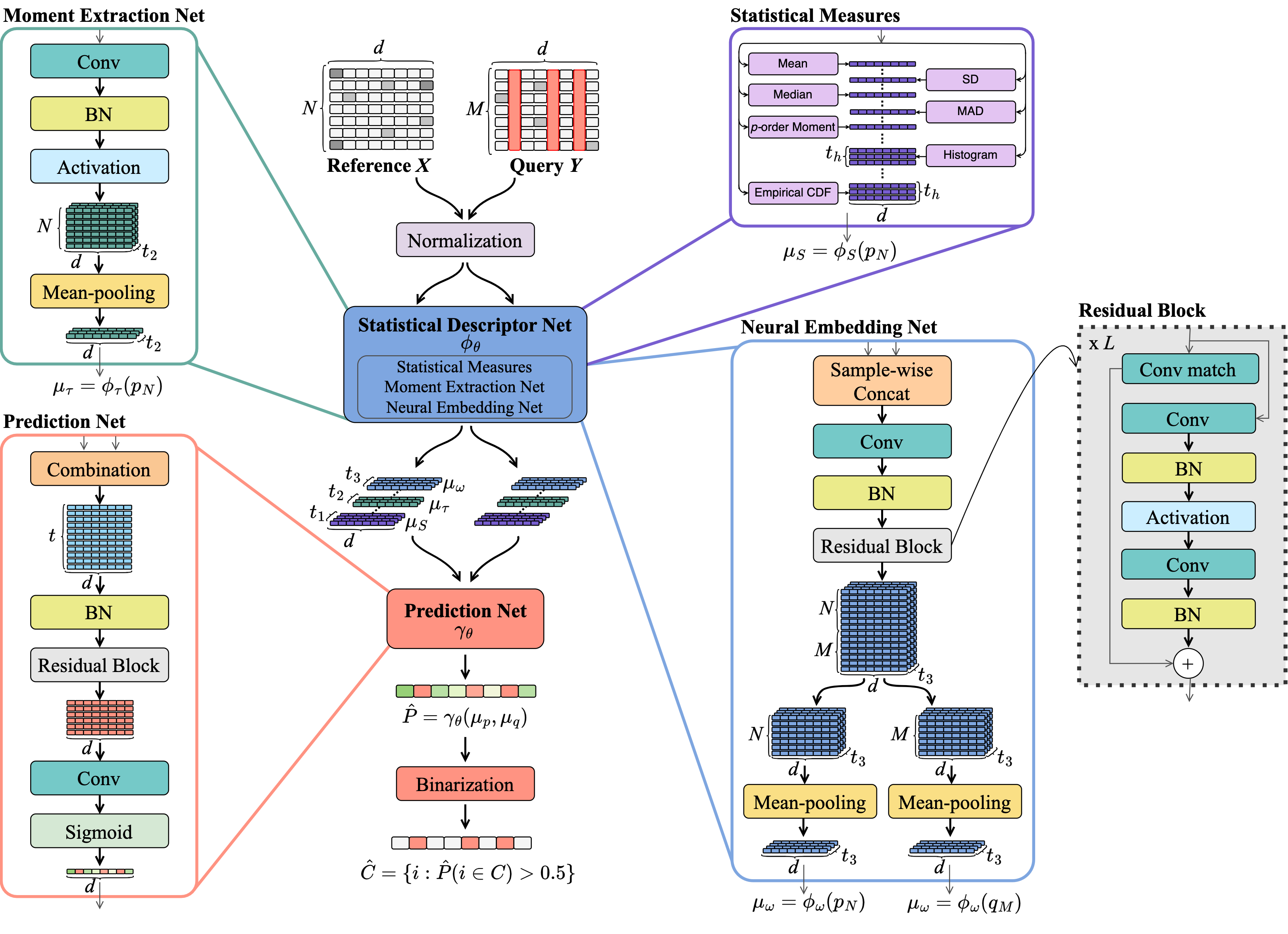}
    \caption{Diagram illustrating the FSL-Net architecture.}
    \label{fig:architecture}
\end{figure*}

\textbf{Problem Formulation.} We follow a similar problem formulation as described in \cite{kulinski2020feature, NEURIPS2023_b3cd64dd}, with minor adaptations:

\textbf{Definition 1. [Empirical Feature Shift Localization Task]} \textit{We are given two sets of $d$-dimensional samples $X = \{x_1, x_2, ..., x_N\}$ and $Y = \{y_1, y_2, ..., y_M\}$ from distributions $p$ and $q$ respectively, with $x_i, y_i \in \mathbb{R}^d$, $|X|=N$, and $|Y|=M$. The feature shift localization task consists of predicting the subset of shifted features $C$ from the input data $X$ and $Y$ using a mapping $F$, such that $C = F(X, Y)$, that satisfies $D(p_{\overline{C}},q_{\overline{C}}) = 0 $, $D(p,q) > 0$, and $C = \argmin_{D(p_{\overline{C}},q_{\overline{C}}) = 0}|C|$, where $D$ is a valid statistical distance.}

A feature shift between $d-$dimensional distributions $p$ and $q$ is present if after removing the \enquote{corrupted} dimensions $C$ and keeping only the \enquote{non-corrupted} dimensions $S=\overline{C}$, with $d = |S| + |C|$, the divergence between the restricted distributions is $D(p_S,q_S) = 0$. The number of corrupted features $|C|$ is assumed to be unknown. We refer to $X$ and $Y$ as the \enquote{reference} and \enquote{query} datasets, and to $p$ and $q$ as the \enquote{reference} and \enquote{query} distributions, respectively. 
In practice, $p$ and $q$ are unknown and only accessible through the samples $X$ and $Y$, requiring the task to be approximated by a method $\tilde{F}$ that maps the input data to the predicted set of corrupted features: $\hat{C} = \tilde{F}(X, Y)$. Such method $\tilde{F}$ can have the form of machine learning-based hypothesis testing \cite{kulinski2020feature}, iterative heuristic algorithms as in DataFix \cite{NEURIPS2023_b3cd64dd}, or the end-to-end trained parametric neural network introduced in this paper. Typically, $\tilde{F}$ is designed to predict a set $\hat{C}$ that is as close as possible to the true set of corrupted features $C$. This set can be represented either as a collection of positional indices or as a $d-$dimensional Boolean vector $C \in \{0,1\}^d$, where each entry indicates whether a corresponding feature is corrupted or not. Note that the vector representation is equivariant to the feature ordering. In this paper, we use the set and vector notations interchangeably unless unclear from the context. We focus on scenarios where either no features or only a subset exhibit a shift. If all features experienced a shift, it would be impossible to determine whether the differences arise naturally between the reference and the query. Thus, we base our approach on the assumption that the true, unmodified query originates from the same distribution as the reference.

As described in \cite{NEURIPS2023_b3cd64dd}, the presented definition of feature shifts covers a wide range of distribution shifts: marginal shifts with $D(p_i, q_i) > 0$, where $p_i$ and $q_i$ represent the marginal distribution of the $i$th dimension; correlation shifts with $D(p, q) > 0$ and $D(p_i, q_i) = 0$ for all $i$, where marginal distributions match but the multi-variate distributions do not; and similarly, correlation shifts with $D(p_S, q_S) = 0$ and $D(p_C, q_C) = 0$ but $D(p, q) > 0$, where correlations are maintained locally, but a shift is present when considering $C$ and $S$ simultaneously. Note that this framework can also model label shifts in regression or classification tasks by simply considering the label as an additional dimension of $p$ and $q$.

\textbf{Feature Shift Localization Network Overview.} 
The proposed Feature Shift Localization Network (FSL-Net) is a model trained end-to-end to predict the set of corrupted features: $\hat{C} = F_\theta(X, Y)$. FSL-Net infers the probability of each feature being part of $C$; that is, the network takes as input the reference and query datasets and predicts a $d-$dimensional vector of probabilities, $\hat{P} = \psi_{\theta}(X,Y)$, such that the $i$th dimension indicates the probability of the $i$th feature being corrupted $\hat{P}(i \in C) = \hat{P}_i = \psi_{\theta}(X, Y)_i$. The complete predicted set of corrupted features $\hat{C}$ can then be obtained by selecting all the features with a probability higher than 0.5: $\hat{C} = \{ i : \hat{P}(i \in C) > 0.5 \}$. 

The Feature Shift Localization Network has two main sub-networks: the Statistical Descriptor Network $\phi_{\theta}$, which generates a finite-dimensional vector summarizing the input distribution $\mu_p = \phi_{\theta}(p)$ (and equivalently for $q$), and the Prediction Network $\gamma_{\theta}$, which takes both vectors and predicts the corruption probability for each feature: $\hat{P} = \gamma_{\theta}(\mu_p, \mu_q)$.
The network is designed to generalize well across datasets of different feature dimensionalities and sample sizes, and by using convolutions, it can scale to high-dimensional and large datasets without requiring re-training.

\textbf{Statistical Functionals and Statistical Functional Maps.} A statistical functional is a mapping $T$ that takes as input a cumulative density function (CDF) $P$ (or similarly a pdf) of a distribution $p$ and outputs a scalar or vector $\mu = T(P) = T(p)$.
Some examples of statistical functionals include the mean, variance, mode, and histograms of the distribution. Because $p$ and $q$ are unknown and only accessible through $X$ and $Y$, we extract the statistical functionals from the empirical distributions $p_N$ and $q_M$, which are constructed by assigning equal probability mass to each of the $N$ and $M$ samples of $X$ and $Y$, respectively. Statistical functionals of interest include linear functionals of the form $\mu = \mathcal{A}(p) = \int g(x) dP(x)$, which can be expressed as a simple average for empirical distributions: $\mu_N = \mathcal{A}(p) = \frac{1}{N} \sum_{j=1}^N g(x_j)$. Note that $g(\cdot)$ does not need to be a linear mapping and can be any (potentially non-linear) function, even a neural network. Empirical mean and histogram estimates are examples of linear functionals.

In this paper, we extend the concept of statistical functionals to \textit{statistical functional maps} -- mappings $T$ that project a $d-$dimensional multivariate CDF $P$ (or equivalently a pdf $p$) into a $d \times t$ tensor $\mu \in \mathbb{R}^{d \times t}$, where the $i$th and $k$th component $\mu_{i,k} = T^k(P, i)$ is obtained by applying the mapping $T^k$ to the multivariate distribution $P$ while using the positional information $i$ (i.e., dimension index). An example of a statistical functional map is a tensor representing a histogram of $t$ bins for each of the $d$ marginal distributions, which can be computed as $\mu^{H}_{i,k} = \frac{1}{N} \sum_{j=1}^N \mathbf{1}_{b_k}(x_{j,i})$, with $1 \leq k \leq t$ and $1 \leq i \leq d$, where $x_{j,i}$ is the $i$th dimension of the $j$th sample, and $\mathbf{1}_{b_k}(x)$ equals $1$ if $x$ is in the interval $b_k$ defining the $k$th bin, and 0 otherwise. Other examples include tensors capturing the first $t$ moments of the $d$ marginal distributions or the $d \times d$ covariance matrix $\mu^{C}_{i,k} = \text{Cov}_p(i,k)$ of distribution $p$, with $t=d$. Statistical functional maps, indexed by dimension $i$, provide finite-dimensional summaries of multivariate distributions, where each $i$th component captures the statistical properties of the $i$th dimension of the distribution and its interactions with other dimensions.

\textbf{Statistical Descriptor Network.} 
The Statistical Descriptor Network is the first component of FSL-Net. This network extracts multiple statistical functional maps from the reference and query datasets, which are then fed into the Prediction Network to localize potential shifts. The Statistical Descriptor Network extracts three statistical functional maps: (1) a non-parametric map $\mu_S = \phi_S(p_N)$, with $\mu_S \in \mathbb{R}^{d \times t_1}$, consisting of simple statistical measures such as marginal means and histograms; (2) a map predicted by a parametric shallow network named Moment Extraction Network $\mu_\tau = \phi_\tau(p_N)$, with $\mu_\tau \in \mathbb{R}^{d \times t_2}$, designed to extract second and higher moments of the data; and (3) a final map extracted by a parametric deep residual network named Neural Embedding Network $\mu_\omega = \phi_\omega(p_N)$, with $\mu_\omega \in \mathbb{R}^{d \times t_3}$, designed to extract a richer representation of the dataset. The maps are concatenated, generating a unique map $\mu_p \in \mathbb{R}^{d \times t}$ describing distribution $p$ (and similarly for $q$), with $t = t_1 + t_2 + t_3$. Note that $d$ varies with each input dataset, but $t$ remains fixed across datasets and only depends on the network hyperparameters. We use $\mu_p = \phi_{\theta}(p_N) = \left[\mu_S; \mu_\tau; \mu_\omega \right]$ to denote the complete mapping of the three statistical functional maps for distribution $p_N$.

\underline{Statistical Measures.}
The first statistical functional map $\mu_S = \phi_S(p_N)$ contains various measures that capture key statistical properties of individual features in the reference and query datasets. Namely, these measures include the mean (indicating the average value), standard deviation (indicating the spread), median (indicating the midpoint), mean absolute deviation (indicating the average distance from the mean), $p-$order moments (capturing univariate higher-order characteristics), marginal histograms (approximating marginal pdfs), and empirical marginal CDFs. Table \ref{tab:stat_measures} presents the formulas for each measure. Note that except for histograms, empirical CDFs, and $p-$order moments, all measures can be represented with arrays of dimensionality $d \times 1$, while the histograms and empirical CDFs are represented in tensors of size $d \times t_h$, where $t_h$ is the number of bins or powers $p$ used. By concatenating all the measures, we obtain a statistical functional map $\mu_S \in \mathbb{R}^{d \times t_1}$.

\begin{table}[htpb]
\scriptsize
\centering
\caption{Statistical measures of the Statistical Descriptor Network. $N$ is the number of observations in $X$, $x_j$ denotes the $i$th dimension of the $j$th sample (subscript $i$ is omitted for brevity), $\epsilon$ is a small positive constant for numerical stability, $b_k$ and $c_k$ are histogram intervals and CDF thresholds, respectively, and $\overline{x}$ and $\sigma$ are the empirical mean and standard deviation of the $i$th dimension.}
\label{tab:stat_measures}
\begin{tabular}{l l l}
\toprule
Statistical Measure & Linear & Equation $\mu_{i,k}$ \\
\midrule
Mean & Yes & \( \overline{x} = \frac{1}{N} \sum_{j=1}^{N} x_{j} \) \\
Standard Deviation & No & \( \sigma = \sqrt{\frac{1}{N} \sum_{j=1}^{N} (x_{j} - \overline{x})^2} \) \\
Median & No & \(  x_{(\frac{N}{2})}  \) \\
Mean Absolute Deviation & Yes & \( \frac{1}{N} \sum_{j=1}^{N} |x_j - \overline{x}| \) \\
\( p \)-order Moments & Yes & \( \frac{1}{N} \sum_{j=1}^{N} x_j^p \) \\
Histogram & Yes & \(  \frac{1}{N} \sum_{j=1}^N \mathbf{1}(x_{j} \in b_k) \) \\
Empirical CDF & Yes &  \(  \frac{1}{N} \sum_{j=1}^N \mathbf{1}(x_{j} \leq c_k) \) \\
\bottomrule
\end{tabular}
\end{table}

\underline{Moment Extraction Network.}
The previously described statistical measures work well for capturing marginal distributions, but they fail to capture correlations and higher-order relations between dimensions. While the covariance matrix could address this, it becomes impractical for high-dimensional datasets due to its quadratic growth with the number of features. In order to capture higher-order relations between features, we make use of the Moment Extraction Network $\mu_\tau = \phi_\tau(p_N) = \frac{1}{N} \sum_{i=1}^N (Wx_i + b)_+$, where $(W,b)$ represent an affine mapping, and $(\cdot)_+$ is the ReLU activation function. In order to adapt to changing input dimensionalities $d$, the affine mapping is parametrized using a convolutional layer and a batch normalization layer. By applying padding, the output of the network has dimensionality $d \times t_2$, where $t_2$ is the number of output channels of the convolution. The convolutional and ReLU layers are applied to each sample $x_j$ independently, and a sample-wise mean pooling operation is applied to obtain a dataset-level vector. Note that this network can be seen as approximating generalized moments of the data \cite{gretton2012kernel, li2015generative, perera2022generative}. Both the Moment Extraction Network and the Neural Embedding Network are trained jointly with the Prediction Network by using a cross-entropy loss and auxiliary loss (see the \textit{loss functions} Section).

\underline{Neural Embedding Network.}
To complement the statistical measures and the Moment Extraction Network, we include a convolutional deep residual network to predict a linear statistical functional map: $\mu_\omega = \phi_\omega(p_N) = \frac{1}{N} \sum_{j=1}^N \omega_{\theta}(x_j)$. 

The Neural Embedding Network, as well as the Prediction Network detailed below, is built upon residual blocks. Each residual block consists of alternating 1D-convolutional layers across features, batch normalization (BN), and a hyperbolic tangent activation function (Tanh). Skip connections are incorporated to facilitate efficient information flow and mitigate vanishing gradients. We conducted experiments with attention-based Multi-Layer Perceptrons (MLPs) using unit-kernel convolutions, but a fully convolutional design worked best (see Section \ref{sec:attention}).

The $i$th output of the statistical functional map predicted by the Neural Embedding Network has the form $\mu_{\omega,i} = \frac{1}{N} \sum_{j=1}^N \left( Wx_{j,i} +  \sum_{h=1}^H \omega_h(x_j) \right)$, where $Wx_{j,i}$ is an affine transformation of the $i$th input feature, and $\omega_h(x_j)$ denotes the output of each of the $H$ residual blocks, capturing non-linear relationships between features.

The complete Statistical Descriptor Network is applied to $X$ and $Y$, yielding $\mu_p = \phi_{\theta}(p_{N}) = \phi_{\theta}(X)$ and $\mu_q = \phi_{\theta}(q_{M}) = \phi_{\theta}(Y)$. Note that $\phi_{\theta}(\cdot)$ is shared between the reference and the query. These descriptors are then fed into the Prediction Network.

\begin{table*}[t]
\scriptsize
\caption{Manipulation types applied to continuous and/or categorical features during training and validation.}

\label{tab:table_train_val_manipulations}
\begin{center}
\begin{tabular}{lp{2.4cm}p{4.8cm}p{3.4cm}l}
\toprule
Type & Mapping & Description & Shift & Data \\
\midrule
T1 & $\beta_i x_i$ \newline $\beta_i \sim \text{Uniform}(0,1)$ & Each value is multiplied by a random number between 0 and 1. & $p_i \neq q_i$ & Cont. \\
T2 & $\beta_i(1 - x) + (1-\beta_i)x$ \newline  $\beta_i \sim \text{Uniform}(0,1)$ & Each value is replaced by a linear combination of $x$ and its negation. & $p_i \neq q_i$ & Cont. \\
T3 & $x_i \sim {p_N}_i$ & The $i$th feature of $Y$ is replaced by sampling from the ref. empirical marginal distribution ${p_N}_i$. & $p_i = q_i, p_C \neq q_C,$ \newline $q_C = \prod_{i \in C} q_i$ & Both\\
T4 & $\text{clamp}_{0,1}(x + \epsilon)$ \newline $\epsilon \sim \text{Normal}(\mu, \sigma)$ & Add Gaussian noise with $\mu \sim \text{Uni.}(-0.2,0.2)$ and $\sigma \sim \text{Uni.}(0.001,0.5)$. & $p_i \neq q_i, \mathbb{E}[q_i] \approx \mathbb{E}[p_i] + \mu$ & Cont. \\
T5 & $\text{CNN}(x)$ & Forward through a CNN with min-max normalization or binarization. & $p_i \neq q_i$ & Both\\
T6 & $x_C \sim {p_N}_C$ & Similar to manipulation (c), but all the features within $C$ are sampled simultaneously from ${p_N}_C$. & $p_i = q_i, p_C = q_C, p \neq q$ & Both\\
T7 & $\text{KNN}(x)$ \newline $K = \{1\!-\!4,7\!-\!9\}$ & Predict feature with KNN (Regressor). & - & Cont.\\
T8 & $\text{KNN}(x)$ \newline $K = \{1\!-\!4,7\!-\!9\}$
 & Predict feature with KNN (Classifier). & - & Cat.\\
\bottomrule
\end{tabular}
\end{center}
\end{table*}

\textbf{Prediction Network.} The Prediction Network $\gamma_{\theta}$ combines the statistical functional maps $\mu_p = \phi_{\theta}(p_{N})$ and $\mu_q = \phi_{\theta}(q_{M})$ to predict a vector of probabilities $\hat{P}=\gamma_\theta(\mu_{p}, \mu_{q})$, indicating the likelihood of each feature belonging to the corrupted set $C$. These maps are combined through an operation $\alpha(\cdot)$, producing a joint statistical map $\mu_{p,q} = \alpha(\mu_{p}, \mu_{q})$. After evaluating various merging operations (see Section \ref{app:extended_merging_operations}), we selected the normalized squared difference: $\alpha(\mu_{p}, \mu_{q}) = \frac{(\mu_{p} - \mu_{q})^2}{||\mu_{p}|| + \epsilon}$, 
where \(\epsilon\) is a small positive constant for numerical stability. This approach enables the network to compare statistical maps in a manner that accounts for their relative magnitudes. The resulting joint representation has dimensions $\mu_{p,q} \in \mathbb{R}^{d \times t}$. The Prediction Network then applies multiple residual blocks to the joint map $\mu_{p,q}$ and produces the final probability estimates through a sigmoid activation layer. Note that, by employing the squared difference, the Prediction Network observes a difference between statistical maps resembling the Maximum Mean Discrepancy (MMD) metric \cite{gretton2012kernel, li2015generative} and acts as a mapping from a distance between distributions into shift probability estimates.

The complete structure of the feature shift localization network has the following form:

\begin{equation}
  \hat{P} = \psi_\theta(p_N, q_M) = \psi_\theta(X, Y) = \gamma_\theta(\phi_\theta(p_N), \phi_\theta(q_M))
\end{equation}

\textbf{Loss Functions.} FSL-Net is trained end-to-end using the binary cross-entropy loss between the predicted probabilities $\hat{P}$ and the ground truth corrupted feature set $C$: $\ell_{CE}(C, \hat{P}) = \sum_{k=1}^d C_k\log(\hat{P}_k) + (1-C_k)\log(1-\hat{P}_k)$. We also add an auxiliary loss function to the predicted statistical functional maps to encourage the learning of useful discriminative features and enforce locality: $\ell_{\text{aux}}(C, \mu_p, \mu_q) = \frac{||{\mu_p}_{\bar{C}} - {\mu_q}_{\bar{C}}||^2}{||{\mu_p}_C - {\mu_q}_C||^2}$. The loss is computed and averaged across multiple labeled datasets with simulated shifts, denoted as $D = \{(C^{(z)}, X^{(z)}, Y^{(z)}) : 1 \leq z \leq N_D\}$, resulting in the total loss function: $\mathcal{L}(\psi_\theta, D) = \sum_{z=1}^{N_d} \ell_{CE} \left( C^{(z)}, \psi_\theta(X^{(z)}, Y^{(z)}) \right) + \lambda \ell_{\text{aux}}(C^{(z)}, \phi_\theta(X^{(z)}), \phi_\theta(Y^{(z)}))$. In practice, $\mathcal{L}(\psi_\theta, D)$ is approximated by using mini-batches and gradient accumulation and optimized with the Adam optimizer.

\textbf{Sample-wise Invariance, Feature-wise Equivariance, and Locality.} Ensuring feature equivariance and sample invariance can help neural networks generalize across datasets with varying numbers of features and samples; (a) Sample-wise invariance: The mean pooling operation used in the linear functionals, Neural Embedding, and Moment Extraction Networks provides maps with shapes that are independent of dataset size, allowing FSL-Net to generalize across datasets of different dimensions. Additionally, the non-linear statistical functionals are computed with the samples sorted from smallest to largest, making them invariant to sample order; (b) Feature-wise invariance: The statistical measures are applied to the marginal distributions, making them equivariant to feature ordering. Convolutions, however, are not typically feature-wise equivariant. Instead, FSL-Net approximates feature invariance by shuffling features in each training mini-batch, enforcing learned representations to be invariant to feature order; (c) Locality: We enforce that the $i$th dimension of the statistical functional maps primarily captures the statistical properties of the $i$th input feature by (1) using marginal distribution-based statistical measures, (2) incorporating a residual connection in the Neural Embedding Network, and (3) applying the auxiliary loss to the statistical functional maps.

\textbf{Training and Validation Datasets.} We source a total of 1,032 diverse tabular datasets from OpenML \cite{van2013openml}, with 10–28k features and 500–3.6M samples, covering continuous, categorical, and mixed data types. Additionally, we generate 368 simulated datasets: 184 based on probabilistic distributions (Gaussian, Bernoulli, and Beta mixtures) and 184 from algebraic functions (Polynomial, Sine, and Logarithmic), each with 5,000 samples and 1,000 features. In total, 1,350 datasets are used for training, with 50 reserved for validation. Section \ref{app:train_val_datasets} describes in detail the dataset selection, preprocessing, and simulation procedures.

\textbf{Training and Validation Manipulations Simulations.} During training, datasets are shuffled in sample and feature order, and random subsets of samples (from 1,000 to 10,000) and features (from 8 to 256) are selected, with each feature normalized to a range of 0 to 1. Each subset is then split equally into reference and query samples. In the query set, a random subset of features (up to 25\%) is manipulated based on feature type (continuous or categorical). Validation batches follow the same process but are limited to 2,048 features. Manipulations, outlined in Table \ref{tab:table_train_val_manipulations} (T1 to T8), are selected with probabilities inversely proportional to the validation F-1 score performance observed during training for each given type. Multiple manipulations are used to simulate a wide range of feature shifts. Note that the manipulations applied during training and validation differ from those applied to the test set.

\section{Experimental Results}

\begin{figure*}[t]
    \centering
    \includegraphics[width=1.0\textwidth]{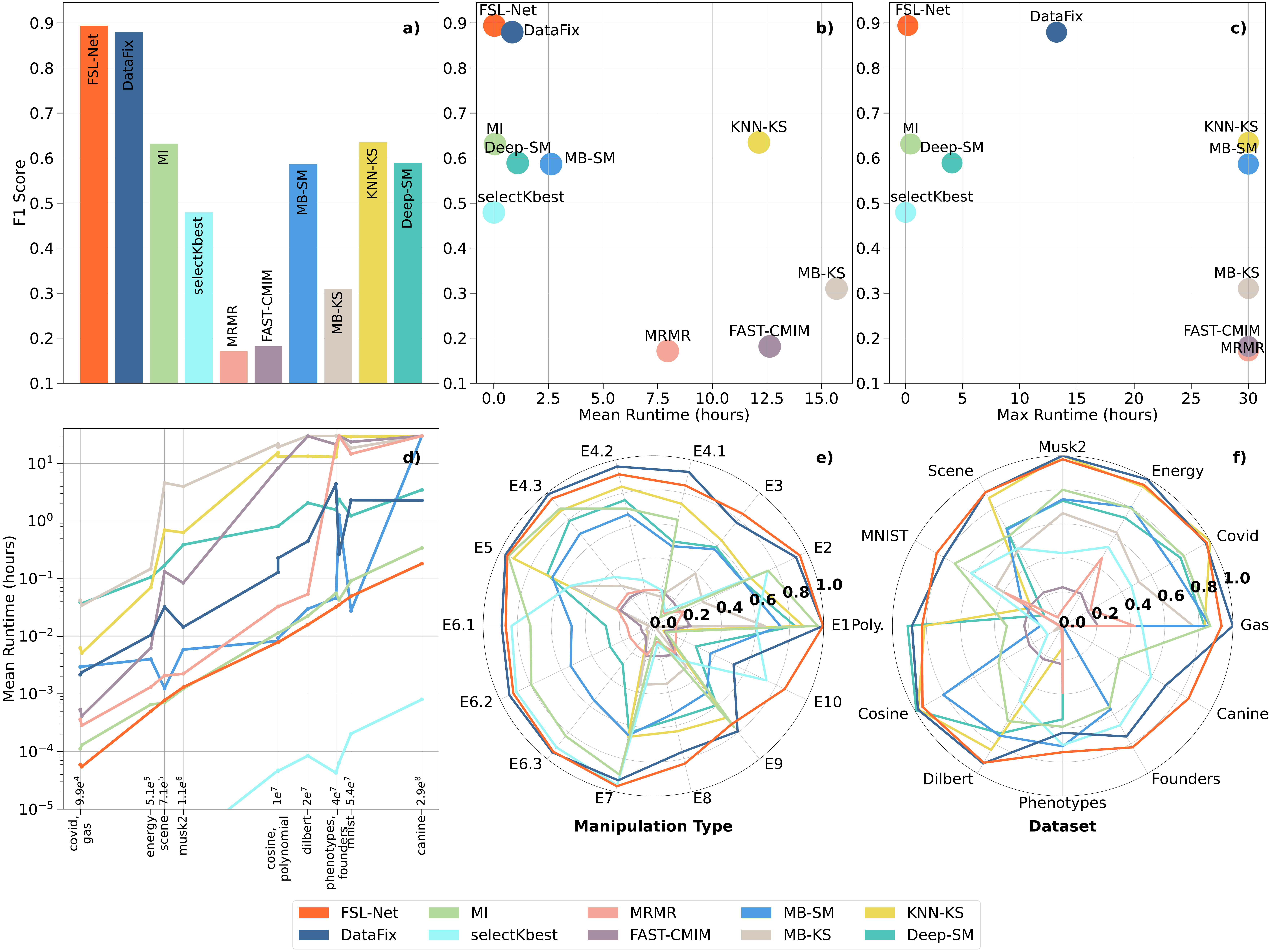}
    \caption{Performance and runtime comparison across feature shift localization methods: a) mean F-1 scores across manipulation types, fractions of manipulated features, and datasets; b) mean F-1 scores vs. mean runtime; c) mean F-1 scores vs. maximum runtime; d) mean runtime vs. sample-feature size product per dataset; e) mean F-1 scores by manipulation type; f) mean F-1 scores by dataset.}
    \label{fig:combined_plot}
\end{figure*}

\textbf{Evaluation Setup.} Our evaluation setup is consistent with \cite{NEURIPS2023_b3cd64dd}, using the same reference and query sets and optimized benchmarking methods. Hyperparameter tuning for FSL-Net is detailed in Section \ref{app:hyper_tuning}. We compare FSL-Net against five feature shift localization methods (DataFix, MB-SM, MB-KS, KNN-KS, and Deep-SM) and four feature selection methods (MI, SelectKBest, MRMR, and Fast-CMIM). The SelectKBest method employs the Chi-square test for categorical datasets and the ANOVA-F test for continuous datasets. For MB-SM, MB-KS, KNN-KS, and Deep-SM, the number of manipulated features $|C|$ is provided, while the other methods, including FSL-Net, do not require it, reflecting a more realistic setting.

\begin{table}[!htpb]
\scriptsize
\centering
\caption{Datasets used during benchmarking.}
\label{tab:benchmark_datasets}
\begin{tabular}{l l l l}
\toprule
Dataset Type & Dataset & \# Features & \# Samples \\
\midrule
\multirow{9}{*}{Continuous} 
    & Gas        & 8       & 12,815  \\
    & Covid      & 10      & 9,889   \\
    & Energy     & 26      & 19,735  \\
    & Musk2      & 166     & 6,598   \\
    & Scene      & 294     & 2,407   \\
    & MNIST      & 784     & 70,000  \\
    & Polynomial & 1,000   & 10,000  \\
    & Cosine     & 1,000   & 10,000  \\
    & Dilbert    & 2,000   & 10,000  \\
\midrule
\multirow{3}{*}{Categorical} 
    & Phenotypes & 1,227   & 31,424  \\
    & Founders   & 10,000  & 4,144   \\
    & Canine     & 198,473 & 1,444   \\
\bottomrule
\end{tabular}
\end{table}

\textbf{Evaluation Data.} We use the same evaluation manipulations and evaluation datasets as in \cite{NEURIPS2023_b3cd64dd}, with the Gas, Covid, and Energy datasets also aligning with those in \cite{kulinski2020feature}. The benchmark datasets vary in data type, feature dimensionality (from 8 to 198,473), and sample size (from 1,444 to 70,000). Table \ref{tab:benchmark_datasets} provides an overview of each dataset, detailing its data type (continuous or categorical), sample size, and feature dimensionality. The continuous datasets are sourced from the UCI repository (Gas \cite{huerta2016online}, Energy \cite{candanedo2017data}, and Musk2 \cite{blake1998uci}) and OpenML (Scene \cite{boutell2004learning}, MNIST \cite{deng2012mnist}, and Dilbert \cite{vanschoren2014openml}). Additionally, a Covid-19 dataset \cite{force2022united} and two simulated datasets generated from algebraic functions that differ from those used in our training and validation simulations (Cosine and Polynomial) \cite{NEURIPS2023_b3cd64dd} are included. The categorical datasets consist of high-dimensional biomedical data, including the Phenotypes dataset \cite{qian2020fast}, a subset of categorical traits from the UK Biobank, the Founders dataset containing binary-coded human DNA sequences \cite{perera2022generative}, and the Canine dataset comprising binary-coded dog DNA sequences \cite{barrabes2023genomic}. Each dataset is normalized on a per-feature basis to a range of 0 to 1, and the samples are evenly divided into two subsets, forming the reference and query sets. As in \cite{NEURIPS2023_b3cd64dd}, a random fraction of query features (5\%, 10\%, or 25\%) undergoes one of 10 manipulation types for continuous data or 8 for categorical data (referred to as manipulations E1-E10). This process generates 30 query sets for continuous data and 24 for categorical data, each with a unique manipulation applied to a given feature subset. Section \ref{app:testing_manipulations} details the evaluation manipulation types. Note that both testing datasets and manipulations differ from those used during FSL-Net training and validation (see Table \ref{tab:table_train_val_manipulations}).

\textbf{Evaluation Protocol and Hardware Specifications.} Performance is evaluated using the F-1 score for feature shift localization accuracy and wall-clock runtime for computational efficiency. Each experiment is run for up to 30 hours, which prevents some methods from completing evaluations on large datasets. To handle incomplete evaluations, we impute missing results with the lowest F-1 score from the same experiment among competing methods and assign them a 30-hour limit. All evaluations were conducted on an Intel Xeon Gold with 12 CPU cores.

\textbf{Feature Shift Localization Performance and Runtime Comparison.} Figure \ref{fig:combined_plot} presents the performance and runtime comparison across different methods. Figure \ref{fig:combined_plot}a presents the average F-1 scores of feature shift localization computed across manipulation types, fractions of manipulated features, and datasets. Figure \ref{fig:combined_plot}b depicts the mean F-1 scores against the mean runtime in hours, while Figure \ref{fig:combined_plot}c illustrates the F-1 scores against the maximum runtime, highlighting the worst-case computational demands. Higher positions in these plots indicate better performance, whereas leftward positions indicate lower runtimes, emphasizing the balance between effectiveness and efficiency. SelectKbest and MI, despite their efficiency, exhibit poor performance in feature shift localization. Methods such as MRMR and FAST-CMIM struggle with scalability and display limited localization capabilities. Notably, FSL-Net achieves the highest average F-1 score, with feature shift localization performance comparable to that of DataFix but at significantly faster speeds. Specifically, FSL-Net is approximately $36\times$ faster on average than DataFix, and up to $136\times$ on the high-dimensional Phenotypes dataset. Additionally, FSL-Net significantly outperforms MB-SM, MB-KS, KNN-KS, and Deep-SM, even though these methods have the advantage of accessing the ground truth $|C|$, while also demonstrating substantially faster computational speeds. While DataFix relies on a computationally intensive iterative optimization heuristic, and MB-SM, MB-KS, KNN-KS, and Deep-SM require dataset-specific training, FSL-Net only requires performing a forward pass through the network for each dataset, providing highly fast inference. 
Figure \ref{fig:combined_plot}d shows the mean runtime as a function of the product between the sample and feature sizes for each dataset, confirming that FSL-Net consistently outperforms DataFix in speed across all datasets, ranking as the second fastest method after selectKBest, comparable to MI, but with much higher accuracy.
Namely, DataFix requires up to 13 hours on the high-dimensional Phenotypes dataset from the UK Biobank and 9 hours on the Canine dataset, whereas FSL-Net completes these tasks in just 2 minutes and 12 minutes, respectively. These results highlight the practical advantage of FSL-Net: it is well-suited for processing high-dimensional large databases, such as those commonly found in biomedicine and e-commerce, making it a valuable alternative to DataFix due to its efficient scalability and excellent localization accuracy. 

\textbf{Performance across Manipulation Types and Datasets.} 
Figure \ref{fig:combined_plot}e and Figure \ref{fig:combined_plot}f show the mean F-1 scores categorized by type of manipulation and dataset, respectively. FSL-Net consistently matches or exceeds the performance of competing methods across all manipulation types, except for E9 and E4, where DataFix exhibits superior performance. These manipulations introduce only minimal perturbations that FSL-Net fails to localize, though lower probability thresholds may improve detection. DataFix fails to accurately detect E10, a shortcoming effectively addressed by FSL-Net. Methods relying on univariate tests, such as MRMR and Fast-CMIM, perform well for manipulations causing marginal distribution shifts but fail entirely with manipulations affecting feature correlations (manipulations E3 and E8). In contrast, techniques based on conditional testing (MB-SM, MB-KS, KNN-KS, and Deep-SM), along with DataFix and FSL-Net, successfully identify these more complex manipulations. FSL-Net shows a clear advantage in high-dimensional datasets (Phenotypes, Founders, and Canine), outperforming all competing methods and highlighting its effectiveness in handling large feature sets. 

\textbf{Ablation Analysis.} We assess the impact of each component of FSL-Net's Statistical Descriptor Network by training models with different combinations of its three components: Statistical Measures (SM), Moment Extraction Network (ME), and Neural Embedding Network (NE). All variants include the Prediction Network (PN), except for the SM-only baseline, where predictions are obtained by directly thresholding the statistical measures. Table \ref{tab:ablation_analysis} presents the mean F-1 score achieved by each configuration. Training lasted three days for all variants, except for the full model, which was trained for an extended period of seven days. Using the Prediction Network in the SM-only baseline greatly improves feature shift localization performance compared to simply thresholding the statistical measures (from an F-1 score of 0.307 to 0.710), emphasizing the crucial importance of the Prediction Network. Each component contributes to performance improvements, suggesting that each component has unique value, with extended training resulting in additional gains.

\begin{table}[htpb]
\scriptsize
\centering
\caption{Mean F-1 scores for various configurations of FSL-Net.}
\label{tab:ablation_analysis}
\begin{tabular}{l l l}
\toprule
Training Duration & Model Configuration & F-1 Score \\
\midrule
\multirow{7}{*}{3-Day} 
& SM & 0.307 \\
& SM + PN  & 0.710 \\
& ME + PN  & 0.742 \\
& NE + PN  & 0.783 \\
& ME + NE + PN  & 0.770 \\
& SM + ME + PN  & 0.855 \\
& SM + NE + PN  & 0.878 \\
& SM + ME + NE + PN  & 0.889 \\
\midrule
\multirow{1}{*}{\textbf{7-Day}} & \textbf{SM + ME + NE + PN} & \textbf{0.894} \\
\bottomrule
\end{tabular}
\end{table}

Additional experimental results are presented in Section~\ref{app:extended_experimental_results}, including: (1) a median-based evaluation of feature shift localization performance and runtime across different methods; (2) a detailed performance and efficiency comparison between FSL-Net and DataFix; (3) extended evaluations of FSL-Net and DataFix on high-dimensional image datasets (CIFAR10 and COIL-100); (4) an analysis of FSL-Net’s runtime improvement over DataFix; (5) an evaluation of the threshold-based variant of the SM-only baseline; and (6) a qualitative evaluation of FSL-Net on the MNIST dataset.

\section{Conclusions}

Current feature shift localization methods involve a trade-off between speed and accuracy when dealing with large data volumes. In this work, we introduced FSL-Net, a novel equivariant neural network that matches or surpasses existing state-of-the-art methods while offering a significant reduction in processing times. FSL-Net leverages neural-learned statistical descriptors, augmented with traditional statistical measures, to effectively capture the input distributions. By contrasting these descriptors across different data sources, the network accurately detects both univariate and multivariate shifts. Most importantly, FSL-Net is designed to manage datasets of varying sizes and to generalize across new distributions without the need for model re-training with each new estimation, providing significant speed and scalability advantages over competing methods.

\section*{Impact Statement}

This paper presents work whose goal is to advance the field
of Machine Learning. There are many potential societal
consequences of our work, none which we feel must be
specifically highlighted here.

\bibliography{paper}
\bibliographystyle{icml2025}

\newpage
\appendix
\onecolumn


\section{Benchmarking Methods}\label{app:benchmarking_methods}

DataFix detects and localizes feature shifts using an iterative adversarial approach called DF-Locate. At each iteration, a random forest classifier is trained to distinguish between samples from a reference and a query distribution. The predicted class probabilities from this discriminator are used to compute the total variation distance (TVD), and its feature importance scores, based on the mean decrease of impurity, are used to locate the features originating the shift. These features are removed in successive rounds until the estimated divergence falls below a threshold or until half the features have been eliminated. A final refinement step uses a knee-detection algorithm to choose the optimal stopping point in the removal process. The selection of features is controlled by a dynamic threshold defined as the product of the TVD and a hyperparameter $\tau$.

Classical univariate statistical filters include Mutual Information statistics (MI) and selectKbest. These methods rank features based on their statistical association with the target. MI measures entropy shared between each feature and the target, while SelectKBest applies ANOVA-F for continuous features and Chi-squared for categorical ones. ANOVA-F tests variance for mean differences in continuous features across target classes, while the Chi-square test evaluates categorical feature-target dependency by comparing observed frequencies across target classes.

More advanced iterative selection methods include Minimum Redundancy Maximum Relevance (MRMR) and Fast Conditional Mutual Information Maximization (FAST-CMIM). These methods attempt to balance feature relevance with redundancy by considering mutual information conditioned on already selected features. MRMR iteratively selects relevant features while minimizing redundancy with the selected features so far. FAST-CMIM iteratively selects features that maximize MI, conditional to previously selected features. These methods, while more expressive than univariate tests, suffer from poor scalability and cannot process large datasets within practical time constraints.

Other benchmarked methods include model-based and statistical testing techniques specifically designed for feature shift localization. These include MB-SM (Multivariate Gaussian with Fisher-divergence test), MB-KS (Multivariate Gaussian with KS test), KNN-KS (K-Nearest Neighbors with KS statistic), and Deep-SM (deep density neural models with Fisher-divergence test). These approaches operate under stronger assumptions about data structure and require prior knowledge of the number of corrupted features, which is rarely available in real-world applications.

\section{Training and Validation Datasets}\label{app:train_val_datasets}

\subsection{OpenML Datasets} 

We source a diverse collection of datasets from OpenML \cite{van2013openml}, selecting only those in a tabular format with at least 10 features and a minimum of 500 samples. To ensure data quality, we preprocess the datasets by removing constant features and addressing missing values using two strategies. In the first strategy, we remove features with more than 40\% missing values, followed by the elimination of samples with missing data. In the second strategy, we remove features with more than 70\% missing values before discarding samples with missing data. If both strategies preserve the required dimensions, we randomly select one for application. We also carefully remove any OpenML dataset that overlaps with the test datasets. For efficiency, large datasets exceeding 40M cell values are partitioned into ten equal-sized sets to accelerate processing. As a result, we obtain a total of 1,032 cleaned and partitioned datasets from OpenML.

\subsection{Algebraic Simulated Datasets} 

We generate 184 simulated datasets using Polynomial, Sine, and Logarithmic functions, each containing 5,000 samples and 1,000 features. Feature values from Polynomial functions are derived from \(f(x) = ax^4 + bx^3 + cx^2 + dx + e\), with parameters \(a\), \(b\), \(c\), \(d\), and \(e\) uniformly sampled from \([-50, 50]\), and the degree randomly set to either 3 or 4. Sine functions follow \(f(x) = a\cdot \cos(bx + c)\), where parameter \(a\) is sampled from \([-50, 50]\), and \(b\) and \(c\) are sampled from \([- \pi, \pi]\). Logarithmic functions are defined as \(\log_{a}(x + 6)\), with base \(a\) sampled from \([2, 10]\). The input values, \(x\), are drawn from the range \([-5, 5]\). Function parameters remain fixed across all samples.

\subsection{Probabilistic Simulated Datasets} 

We generate 184 simulated datasets based on probabilistic distributions, including Gaussian, Bernoulli, and Beta mixture models. Each dataset consists of 5,000 samples and 1,000 features. To construct these datasets, we begin by selecting a base distribution \(\mathcal{D}\) from \{Gaussian, Bernoulli, Beta\} and initializing parameters for a mixture model with a randomly determined number of components \(K\), where \(1 \leq K \leq 100\). For the Gaussian distribution, the mean vector \(\mu_k = (\mu_{k1}, \mu_{k2}, \ldots, \mu_{kd})\) is drawn from a standard normal distribution \(\mathcal{N}(0, 1)\), while the variance vector \(\sigma_k^2 = (\sigma_{k1}^2, \sigma_{k2}^2, \ldots, \sigma_{kd}^2)\) is sampled from a Uniform distribution \(\mathcal{U}(0.1, 1.1)\) for each feature independently. For the Bernoulli distribution, the probability vector \(p_k = (p_{k1}, p_{k2}, \ldots, p_{kd})\), which represents the probability of success for each feature, is sampled from a uniform distribution \(\mathcal{U}(0, 1)\). For the Beta distribution, the parameters \(\alpha_k = (\alpha_{k1}, \alpha_{k2}, \ldots, \alpha_{kd})\) and \(\beta_k = (\beta_{k1}, \beta_{k2}, \ldots, \beta_{kd})\) are drawn from a uniform distribution \(\mathcal{U}(1, 2)\). Samples are generated using the weighted sum of these mixture components, defined as \(\sum_{k=1}^K \pi_k \mathcal{D}(\theta_k)\), where \(\pi_k\) represents the mixing coefficients, satisfying \(\sum_{k=1}^K \pi_k = 1\). 

Each dataset has a 25\% chance of undergoing multiple transformations, with the number of transformation steps \(I\) randomly chosen between 1 and 5. These transformations include normalization (standard or min-max scaling), followed by a linear transformation in which the dataset is multiplied by a randomly sampled matrix \(W \in \mathbb{R}^{d \times d}\). The entries of \(W\) are drawn from one of four distributions: \(\mathcal{U}(0,1)\), \(\mathcal{N}(0,1)\), \(\text{Beta}(1,1)\), or \(\text{Bernoulli}(0.5)\). Additionally, a non-linear transformation is applied, randomly chosen from the ReLU, GELU, sigmoid, hyperbolic tangent, or logarithmic transformation. To preserve structural information, residual connections are sometimes applied, where the transformed data is combined with its previous or initial state, either directly or after undergoing normalization.

\newpage

\section{Testing Manipulation}\label{app:testing_manipulations}

Table \ref{tab:table_benchmark_manipulations} outlines the manipulation types used to induce shifts during evaluation. Some distort marginal distributions (E1, E2, E4, E5, E6, and E7), with manipulation E4 leaving the mean approximately unchanged. Others shuffle feature values across samples, altering feature correlations but not marginal distributions (E3 and E8). Manipulations E9 and E10 use KNN predictions to replace corrupted continuous and categorical features, respectively. For a more detailed explanation, see \cite{NEURIPS2023_b3cd64dd}.

\begin{table}[htpb]
\scriptsize
\centering
\caption{Manipulation types applied to continuous and/or categorical features during benchmarking (Table from \cite{NEURIPS2023_b3cd64dd}).}
\label{tab:table_benchmark_manipulations}
\begin{tabular}{lp{2.4cm}p{4.8cm}p{3.4cm}l}
\toprule
Type & Mapping & Description & Shift & Data \\
\midrule
E1 &  $x \sim \text{Uniform}(0,1)$ & Each value is substituted by a random number between 0 and 1. & $p_i \neq q_i$ & Cont. \\
E2 & $1 - x$ & Each value is negated. & $p_i \neq q_i, \mathbb{E}[q_i] = 1 - \mathbb{E}[p_i]$ & Both \\
E3 & $P_iX_i$ & $P_i$ is a random permutation matrix applied to feature $i$. & $p_i = q_i, p_C \neq q_C,$ \newline $q_C = \prod_{i \in C} q_i$ & Both \\
E4.1-4.3 & $\text{clamp}_{0,1}(x + \alpha \sigma)$ \newline $\sigma \sim \text{Rademacher}(0.5)$ & Add constant noise with a random sign. $\alpha \in \{0.02, 0.05, 0.1\}$ for 4.1-4.3 respectively. & $p_i \neq q_i, \mathbb{E}[p_i] \approx \mathbb{E}[q_i]$ & Cont. \\

E5 & 
$\text{round}(x)$ & Values are binarized. & $p_i \neq q_i$ & Cont.\\
E61-6.3 & $b(1-x) + (1-b)x$ \newline $b \sim \text{Bernoulli}(\rho)$   & Values are negated with probability $\rho \in \{0.2, 0.4, 0.6\}$ for 6.1-6.3 respectively. & $p_i \neq q_i,$ \newline $\mathbb{E}[q_i] = \rho + (1-2\rho)\mathbb{E}[p_i]$ & Cat.\\

E7 &  $\text{MLP}(x)$ & Forward through an MLP with min-max normalization or binarization. & $p_i \neq q_i$ & Both\\
E8 & $PX_i$ & $P$ is a random permutation matrix applied to all features simultaneously. & $p_i = q_i, p_C = q_C, p \neq q$  & Both\\
E9 & $\text{KNN}(x)$ & Predict feature with KNN (Regressor). & - & Cont.\\
E10 & $\text{KNN}(x)$ & Predict feature with KNN (Classifier). & - & Cat.\\
\bottomrule
\end{tabular}
\end{table}

\newpage

\section{Exploration of Attention Mechanisms}
\label{sec:attention}

We investigated the integration of attention mechanisms across features within the residual blocks of both the Neural Embedding Network and the Prediction Network to effectively capture diverse data patterns and feature correlations. Upon incorporation, the attention mechanisms were coupled with MLPs implemented using convolutional layers with unit-sized kernels.

\subsection{Efficient EVA Attention Layers}\label{sec:eva_attention}

Using full attention across features was infeasible for high-dimensional datasets due to its quadratic computational requirements. Consequently, we investigated the use of EVA attention layers \cite{zheng2023efficient}. Attention layers without positional embeddings are inherently equivariant; therefore, we did not include positional embeddings. However, efficient (approximate) attention layers, such as EVA, can fail to preserve equivariance under certain conditions.

\subsection{Sequence Handling and Window Size Adjustment}
\label{sec:sequence_handling}

Key considerations involve ensuring that the input size to the EVA layer is not only smaller than the sequence length but also divisible by it. Managing variable-length sequences requires meticulous data processing, achieved through padding and the application of a key padding mask for efficient sequence handling. The window size $w$ of the EVA attention was dynamically adjusted based on the sequence length to optimize processing efficiency and effectiveness, specifically determined as $w = \min\left(\max\left(8, \left\lfloor\frac{d}{4}\right\rfloor\right), 32\right)$.

\subsection{Limitations}
\label{sec:limitations_alternatives}

Despite EVA attention layers being more efficient than traditional attention mechanisms \cite{liu2021multi}, their integration into both the Neural Embedding Network and the Prediction Network resulted in CUDA out-of-memory issues for certain configurations. Consequently, we attempted to incorporate attention solely within the Prediction Network. While this approach alleviated memory issues, the performance remained below that of our baseline configuration without any attention mechanisms.

\newpage
\section{Hyperparameter Tuning}\label{app:hyper_tuning}

\subsection{Training Setup and Hardware Specifications} 

We conducted a random search across various hyperparameters of the FSL-Net architecture and its training strategy. Training for each network was terminated if the validation loss did not improve for 50 consecutive evaluation intervals, with the maximum training duration limited to three days. To expedite the training process, a single NVIDIA-SMI GPU with 32GB of memory was used. The networks were validated on 50 validation datasets every 2,500 training iterations. The model checkpoint achieving the highest validation accuracy was selected as the optimal network and subsequently trained for up to seven days. This best-performing model was employed for inference and evaluation in the benchmarking experiments.

\subsection{Alternative Statistical Measures}\label{app:extended_statistical_measures}

In addition to the statistical measures delineated in the main text, we investigated the following metrics: skewness (indicating asymmetry), kurtosis (reflecting tail heaviness), index of dispersion (representing the variance-to-mean ratio), and trimmed mean deviation (a robust estimator of central tendency that excludes outliers). The formulas for each measure are presented in Table \ref{tab:stat_measures_appendix}. These measures were ultimately excluded from the final analysis, as they did not confer any discernible advantage.

\begin{table}[htpb]
\scriptsize
\centering
\caption{Formulas for additional statistical measures of the Statistical Descriptor Network. $N$ is the number of observations in $X$, $x_j$ represents the $i$th dimension of the $j$th sample (we skip the subscript $i$ for brevity), $x_{(l)}$ is the $i$th dimension of the $l$th sample after sorting the samples from smallest to largest, $r$ is the count of observations trimmed from each end in trimmed mean deviation, and $\overline{x}$ and $\sigma$ are the empirical mean and standard deviation of the $i$th dimension.}
\label{tab:stat_measures_appendix}
\begin{tabular}{l l l}
\toprule
Statistical Measure & Linear & Equation $\mu_{i,k}$ \\
\midrule
Skewness & Yes & \( \frac{1}{N} \sum_{j=1}^{N} \left(\frac{x_j - \overline{x}}{\sigma}\right)^3 \) \\
Kurtosis & Yes & \( \frac{1}{N} \sum_{j=1}^{N} \left(\frac{x_j - \overline{x}}{\sigma}\right)^4 \) \\
Index of Dispersion & Yes & \( \frac{\sigma^2}{\overline{x} + \epsilon} \) \\
Trimmed Mean Deviation & No & \( \frac{1}{N-r} \sum_{j=r+1}^{N-r} x_{(j)} \) \\
\bottomrule
\end{tabular}
\end{table}

\subsection{Alternative Merging Operations}\label{app:extended_merging_operations}

In addition to the normalized squared difference discussed in the main text, we examined several alternative merging operations within the Prediction Network. The following methods were evaluated; however, none yielded performance improvements over the existing approach: concatenation $\alpha(\mu_{p}, \mu_{q}) = \left[ \mu_{p}; \mu_{q} \right]$; element-wise difference $\alpha(\mu_{p}, \mu_{q}) = \mu_{p} - \mu_{q}$; and squared difference $\alpha(\mu_{p}, \mu_{q}) = (\mu_{p} - \mu_{q})^2$. Note that, for the concatenation method, the merged feature vector has dimensions $\mu_{p,q} \in \mathbb{R}^{d \times 2t}$, whereas for the other merging operations, it has dimensions $\mu_{p,q} \in \mathbb{R}^{d \times t}$.

\subsection{Hyperparameter Search}

\subsubsection{Network Tuning Parameters}

Table \ref{tab:network_tuning_parameters} provides an overview of the search spaces and the optimal values determined for each network-related hyperparameter in FSL-Net. These parameters encompass configurations for the statistical measures, Moment Extraction Network, Neural Embedding Network, and Prediction Network. For parameters not explicitly specified, default values were applied.

\begin{table}[htpb]
\scriptsize
\centering
\caption{Search space and optimal values for tuned network hyperparameters in FSL-Net.}
\begin{tabular}{p{3cm}p{3cm}p{4.3cm}p{2.1cm}}
\toprule
Component & Hyperparameter & Possible Values & Optimal Value \\
\midrule
\textbf{Statistical Measures} 
 & Mean & $\{\text{True}, \text{False}\}$ & \textbf{True} \\
 & Standard Deviation & $\{\text{True}, \text{False}\}$ & \textbf{True} \\
 & Median & $\{\text{True}, \text{False}\}$ & \textbf{True} \\
 & Mean Absolute Deviation & $\{\text{True}, \text{False}\}$ & \textbf{True} \\
 & $p$-order Moments & $\{\text{True}, \text{False}\}$ & \textbf{True} \\
 & $p$ & $\{2\}$, $\{3\}$, $\{2, 3\}$ & \textbf{2, 3} \\
 & Histogram & $\{\text{True}, \text{False}\}$ & \textbf{True} \\
 & Empirical CDF & $\{\text{True}, \text{False}\}$ & \textbf{True} \\
 & \# Bins & $\{100\}$, $\{50, 100\}$ & \textbf{100} \\
 & Skewness & $\{\text{True}, \text{False}\}$ & \textbf{False} \\
 & Kurtosis & $\{\text{True}, \text{False}\}$ & \textbf{False} \\
 & Index of Dispersion & $\{\text{True}, \text{False}\}$ & \textbf{False} \\
 & Trimmed Mean Deviation & $\{\text{True}, \text{False}\}$ & \textbf{False} \\
 & Trimmed Percentage & $\{\text{0.1}\}$ & \textbf{0.1} \\
\midrule
\textbf{Moment Extraction Network} 
 & \# Kernels & $\{32, 64, 128\}$ & \textbf{64} \\
 & Kernel Size & $\{75, 125\}$ & \textbf{75} \\
 & Dilation & $\{1\}$ & \textbf{1} \\
 & Activation & $\{\text{ReLU}, \text{Tanh}\}$ & \textbf{ReLU} \\
\midrule
\textbf{Neural Embedding Network} 
 & \# Residual Layers & $\{3, 5, 7\}$ & \textbf{5} \\
 & \# Kernels & $\{32, 64\}$ & \textbf{64} \\
 & Kernel Size & $\{5, 7\}$ & \textbf{5} \\
 & Dilation & $\{1\}$ & \textbf{1} \\
 & Activation & $\{\text{Tanh}, \text{GELU}\}$ & \textbf{Tanh} \\
\midrule
\textbf{Prediction Network} 
 & \# Residual Layers & $\{3, 5, 7\}$ & \textbf{7} \\
 & \# Kernels & $\{32, 64\}$ & \textbf{64} \\
 & Kernel Size & $\{5, 7\}$ & \textbf{5} \\
 & Dilation & $\{1\}$ & \textbf{1} \\
 & Activation & $\{\text{Tanh}, \text{GELU}\}$ & \textbf{Tanh} \\
 & Combination & \{Concatenation, Element-wise difference, Squared difference, Normalized squared difference\} & \textbf{Normalized squared difference} \\
\bottomrule
\end{tabular}
\label{tab:network_tuning_parameters}
\end{table}

\subsubsection{Attention Tuning Parameters}

Table \ref{tab:attention_tuning_parameters} presents the hyperparameters of the attention mechanism explored during the optimization process. These parameters regulate the EVA attention layers within FSL-Net, including the number of heads, the number of landmarks, the window factor, and dropout rates.

\begin{table}[htpb]
\scriptsize
\centering
\caption{Search space and optimal values for tuned attention hyperparameters.}
\begin{tabular}{p{3cm}p{3cm}p{4.3cm}p{2.1cm}}
\toprule
Component & Hyperparameter & Possible Values & Optimal Value \\
\midrule
\textbf{Attention} 
 & \# Heads & $\{4, 8\}$ & \textbf{4} \\
 & \# Landmarks & $\{8\}$ & \textbf{8} \\
 & Window factor & $\{4\}$ & \textbf{4} \\
 & Window size & $\{7, 9\}$ & \textbf{7} \\
 & Overlapping windows & $\{\text{True}, \text{False}\}$ & \textbf{False} \\
 & Attention dropout & $\{0, 0.3, 0.5\}$ & \textbf{0} \\
 & Projection dropout & $\{0\}$ & \textbf{0} \\
\bottomrule
\end{tabular}
\label{tab:attention_tuning_parameters}
\end{table}

\subsubsection{Optimization Tuning Parameters}

Table \ref{tab:optimization_tuning_parameters} outlines the search space and optimal values for optimization hyperparameters in FSL-Net's training strategy. This includes the loss function and Adam optimizer settings.

\begin{table}[htpb]
\scriptsize
\centering
\caption{Search space and optimal values for optimization hyperparameters in the training strategy.}
\begin{tabular}{p{3cm}p{3cm}p{4.3cm}p{2.1cm}}
\toprule
Component & Hyperparameter & Possible Values & Optimal Value \\
\midrule
\textbf{Loss Function} 
 & $\lambda$ & $\{0.0001, 0.001\}$ & \textbf{0.001} \\
\midrule
\textbf{Adam Optimizer} 
 & Learning Rate & $\{0.0001, 0.001, 0.01\}$ & \textbf{0.001} \\
 & Learning Rate Gamma & $\{0.9, 0.999, 0.9995\}$ & \textbf{0.9995} \\
\bottomrule
\end{tabular}
\label{tab:optimization_tuning_parameters}
\end{table}

\newpage

\newpage

\section{Extended Experimental Results}\label{app:extended_experimental_results}

\subsection{Median-based Feature Shift Localization Performance and Runtime Comparison}

\begin{figure*}[htpb]
    \centering
    \includegraphics[width=1.0\textwidth]{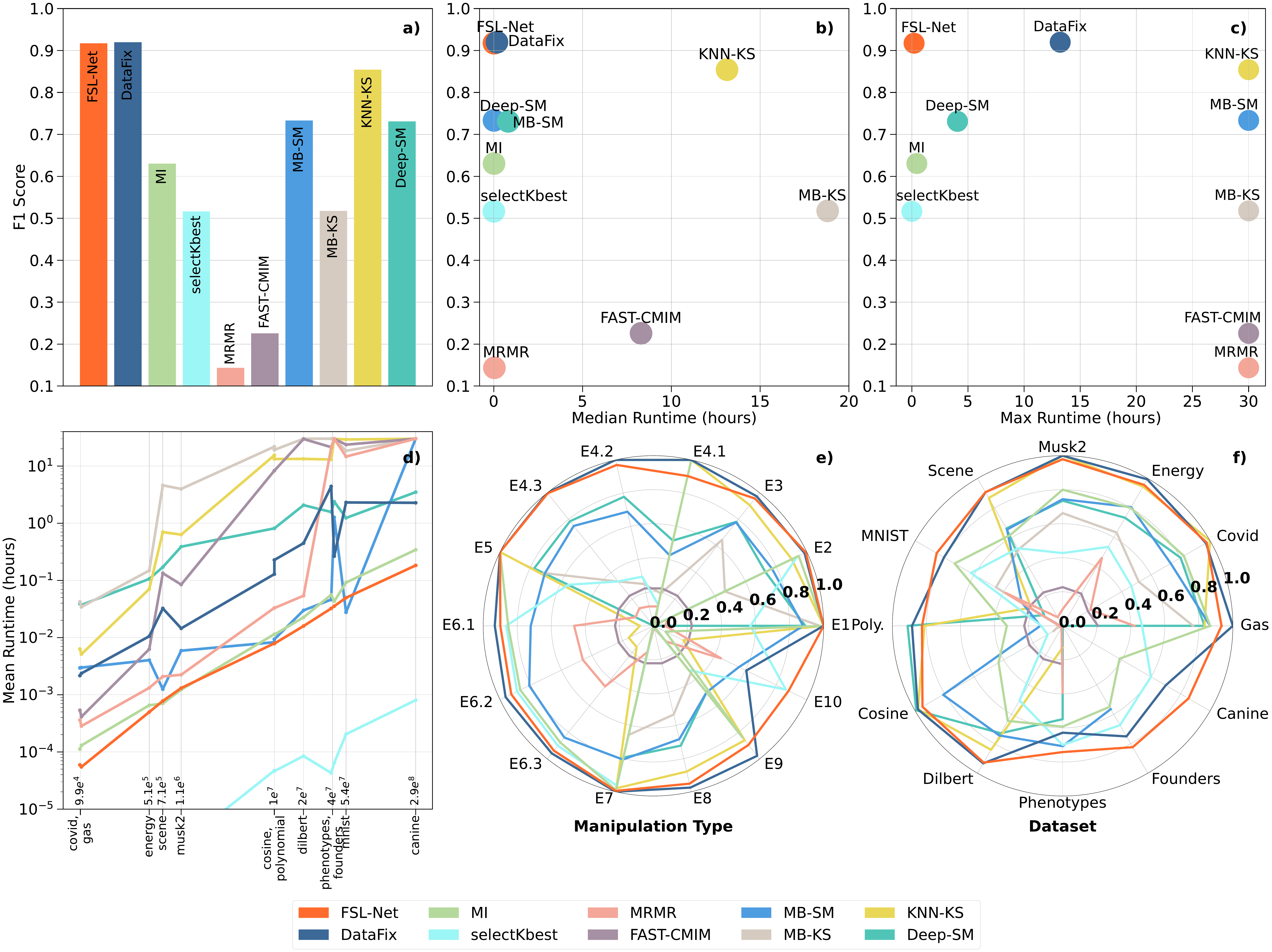}
    \caption{Performance and runtime comparison across feature shift localization methods: a) median F-1 scores across manipulation types, fractions of manipulated features, and datasets; b) median F-1 scores vs. median runtime; c) median F-1 scores vs. maximum runtime; d) mean runtime vs. sample-feature size product per dataset; e) median F-1 scores by manipulation type; f) mean F-1 scores by dataset.}
    \label{fig:median_combined_plot}
\end{figure*}

Figure~\ref{fig:median_combined_plot} presents a median-based analysis of feature shift localization performance and runtime, complementing the mean-based evaluation in the main text. For each method, F-1 scores and runtimes are first averaged across manipulation types and fractions of manipulated features, then aggregated using the median across datasets. This approach eliminates the need for imputation in cases where slower methods fail to complete for some datasets, enabling fairer comparisons across a broader set of methods. However, the mean-based results remain more representative for consistently successful approaches like FSL-Net and DataFix, which completed all evaluations. Figure \ref{fig:median_combined_plot}a summarizes the resulting median F-1 scores, while Figure \ref{fig:median_combined_plot}b and Figure \ref{fig:median_combined_plot}c visualize the trade-off between performance and computational cost, plotting median F-1 scores against median and maximum runtime, respectively. In these plots, higher positions indicate better localization accuracy, while positions further to the left denote lower runtime. These comparisons again confirm that FSL-Net matches DataFix in localization accuracy while offering a substantial advantage in runtime efficiency. Despite being only slightly slower than SelectKBest, FSL-Net far exceeds this method in accuracy. Meanwhile, methods such as MRMR and FAST-CMIM continue to struggle with scalability and performance under median-based evaluation. MB-SM, MB-KS, KNN-KS, and Deep-SM show higher F-1 scores when evaluated by the median rather than the mean, with KNN-KS performing the best among them. However, all of these methods still lag behind FSL-Net, despite benefiting from access to the ground truth $|C|$ -- a condition rarely met in practice -- and suffer from extremely poor scalability on large datasets. Figure \ref{fig:median_combined_plot}d plots the mean runtime against the product of sample and feature sizes for each dataset. FSL-Net maintains a significantly lower mean runtime than DataFix across datasets of varying dimensionality. Figure \ref{fig:median_combined_plot}e presents median F-1 scores by manipulation type, while Figure \ref{fig:median_combined_plot}f shows mean F-1 scores by dataset. FSL-Net matches or surpasses the performance of competing methods across nearly all manipulation types, with two exceptions: E9, where DataFix achieves higher performance, consistent with mean-based results, and E4, where both DataFix and KNN-KS exhibit superior accuracy. Notably, FSL-Net demonstrates a clear advantage on high-dimensional datasets such as Phenotypes, Founders, and Canine, outperforming all baselines and highlighting its effectiveness in scenarios involving large feature sets.

\newpage

\subsection{Performance and Computational Efficiency of FSL-Net and DataFix}

\begin{figure}[htpb]
    \centering
    \includegraphics[width=1.0\textwidth]{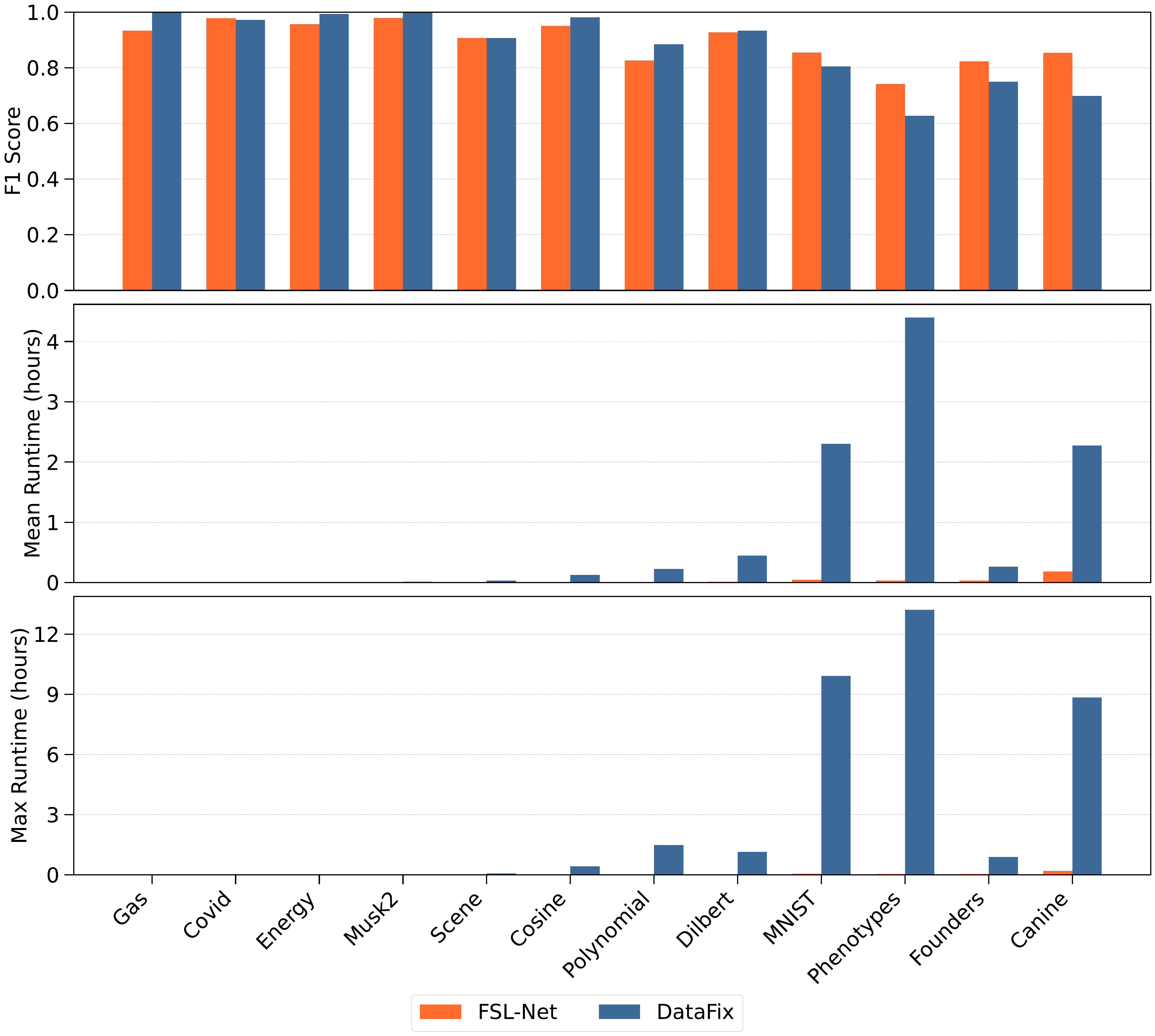}
    \caption{Mean F-1 scores (top), mean runtimes (middle), and maximum runtimes (bottom) of FSL-Net and DataFix by dataset.}
    \label{fig:datafix_vs_fsl}
\end{figure}

Figure \ref{fig:datafix_vs_fsl} displays the mean F-1 scores (top), mean runtimes (middle), and maximum runtimes (bottom) for both FSL-Net and DataFix, computed across various manipulation types and fractions of manipulated features. FSL-Net exhibits comparable feature shift localization performance to DataFix on the majority of datasets, with a notable advantage on large datasets such as Phenotypes, Founders, and Canine, while also demonstrating significantly greater computational efficiency. The pronounced disparity between the mean and maximum runtimes of DataFix on certain datasets suggests that its processing time is highly sensitive to the complexity of the shifts, potentially necessitating multiple iterations to achieve convergence. In contrast, FSL-Net executes only a single forward pass through the network, ensuring scalability to both high-dimensional and large datasets.

\newpage

\subsection{Extended Evaluation of FSL-Net and DataFix on CIFAR10 and COIL-100}

Table~\ref{tab:extended_datasets} presents an extended evaluation of FSL-Net and DataFix on two image datasets: CIFAR10 (10k samples) and COIL-100. We report the mean F-1 score in feature shift localization and the mean and maximum runtime (in hours), computed across manipulation types and fractions of manipulated features. FSL-Net achieves higher F-1 scores on average and consistently outperforms DataFix in terms of computational efficiency, exhibiting significantly lower runtimes in both average and worst-case scenarios.

\begin{table}[!htpb]
\scriptsize
\centering
\caption{Comparison of FSL-Net and DataFix averaged across manipulation types and fractions of manipulated features on two datasets: CIFAR10 (10k) and COIL-100.}
\label{tab:extended_datasets}
\begin{tabular}{lcccccc}
\toprule
\textbf{Dataset} & \multicolumn{2}{c}{\textbf{F-1 Score}} & \multicolumn{2}{c}{\textbf{Mean Runtime (hours)}} & \multicolumn{2}{c}{\textbf{Max Runtime (hours)}} \\
\cmidrule(lr){2-3} \cmidrule(lr){4-5} \cmidrule(lr){6-7}
& \textbf{FSL-Net} & \textbf{DataFix} & \textbf{FSL-Net} & \textbf{DataFix} & \textbf{FSL-Net} & \textbf{DataFix} \\
\midrule
\textbf{CIFAR10} & \textbf{0.9565} & 0.8911 & \textbf{0.0319} & 0.2948 & \textbf{0.0336} & 0.8084 \\
\textbf{COIL-100} & 0.9720 & \textbf{0.9805} & \textbf{0.4906} & 2.9228 & \textbf{0.8406} & 9.1567 \\
\bottomrule
\end{tabular}
\end{table}

\newpage

\subsection{Runtime Improvement of FSL-Net over DataFix}

\begin{figure}[htpb]
    \centering
    \includegraphics[width=1.0\textwidth]{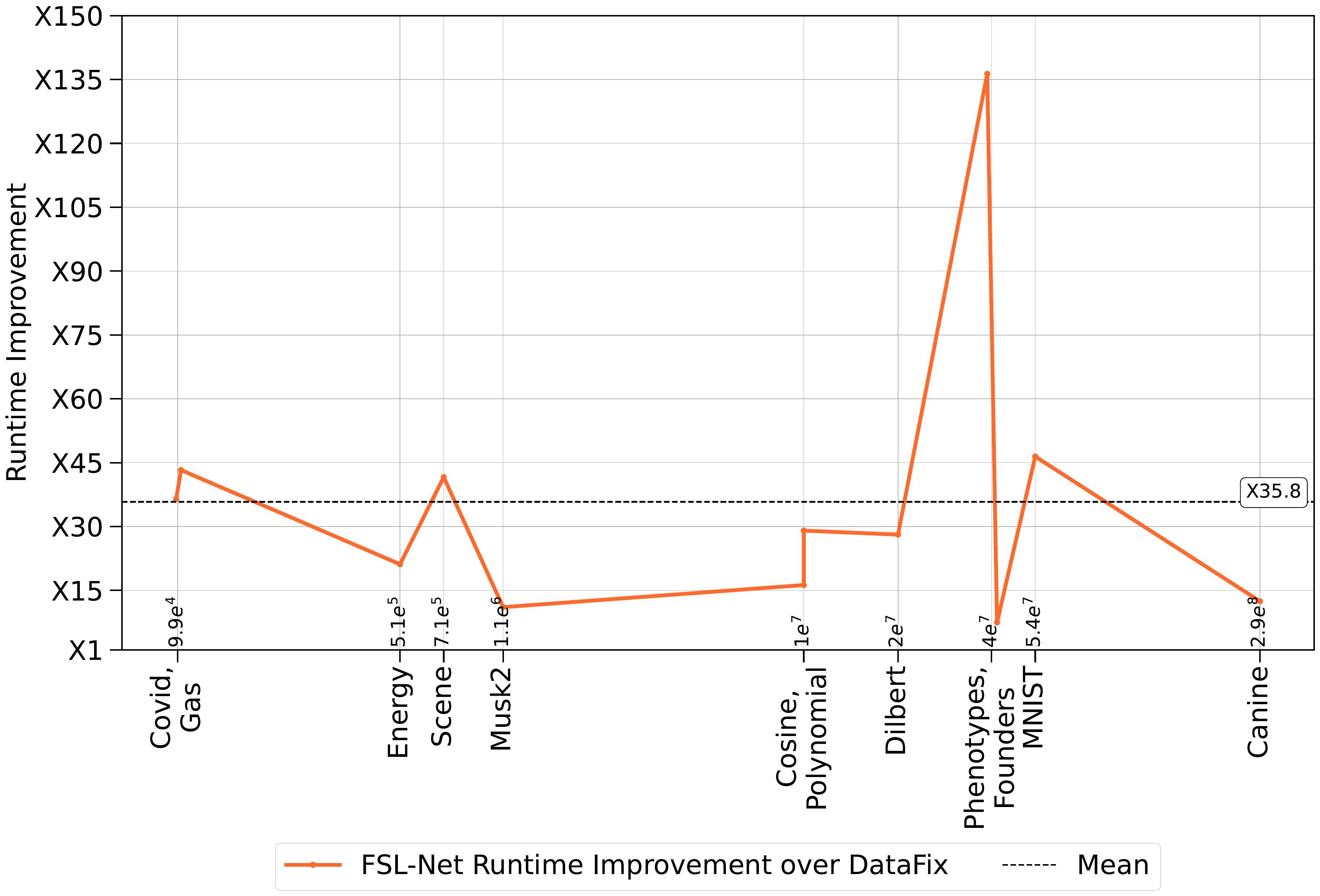}
    \caption{Runtime improvement of FSL-Net over DataFix across datasets, sorted by dataset size (number of samples $\times$ number of features). The speedup factor indicates how many times faster FSL-Net is compared to DataFix. The dashed line marks the average speedup of $35.8\times$.}
    \label{fig:runtime_improvement}
\end{figure}

Figure~\ref{fig:runtime_improvement} presents the mean runtime improvement of FSL-Net over DataFix across datasets, sorted by increasing dataset size (measured as the product of the number of samples and the number of features). The speedup factor -- defined as the ratio of DataFix's runtime to FSL-Net's -- measures the performance gain, with higher values indicating greater efficiency. The results are averaged across manipulation types and fractions of manipulated features. On average, FSL-Net achieves a substantial speedup of $35.8\times$ (indicated by the dashed line), consistently outperforming DataFix across datasets. Among all datasets, the high-dimensional Phenotypes dataset exhibits the greatest speedup, with FSL-Net outperforming DataFix by a remarkable $136.3\times$.

\newpage

\subsection{Performance of the Threshold-Based Variant of the SM-only Baseline}\label{app:sm_thresholding}

\begin{figure}[htpb]
    \centering
    \includegraphics[width=1.0\textwidth]{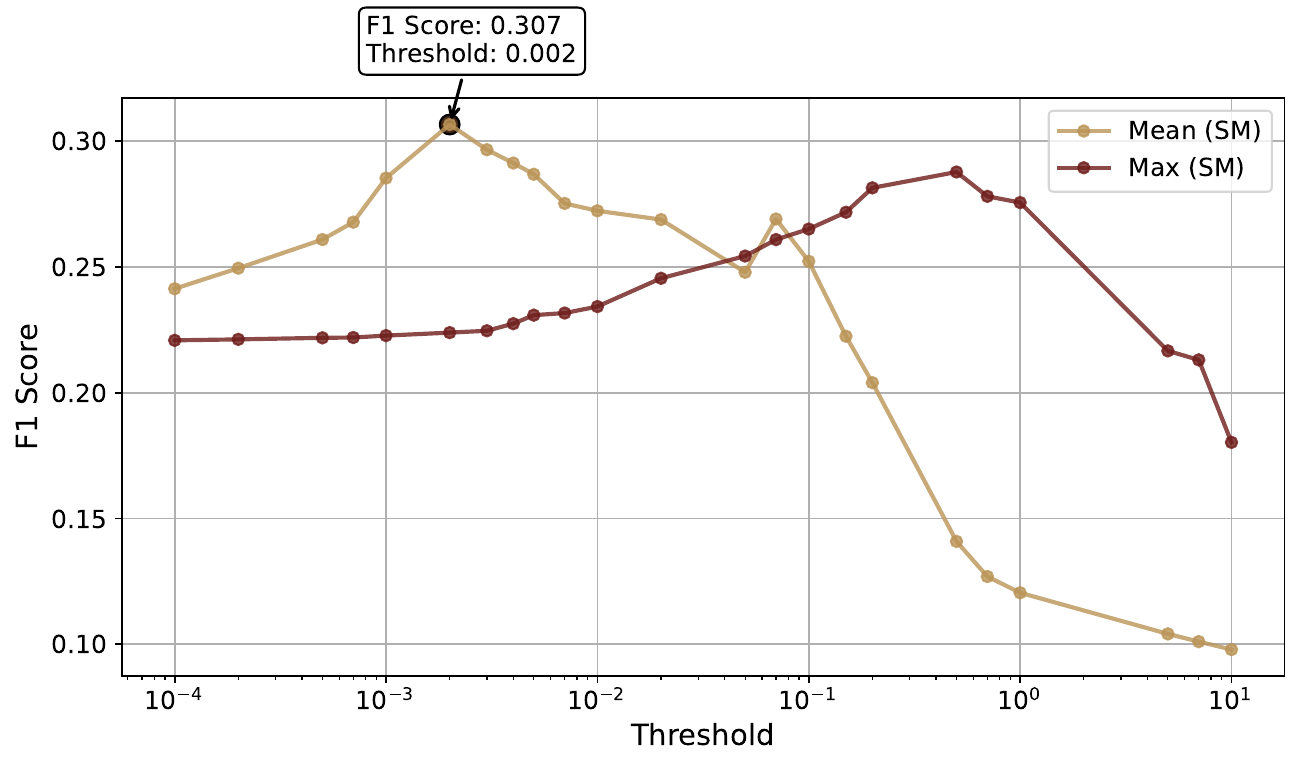}
    \caption{Mean F-1 scores of the SM-only variant of FSL-Net, where the Prediction Network is omitted and feature shift localization relies solely on thresholding the statistical measures. The F-1 scores are averaged across datasets, manipulation types, and fractions of manipulated features. Statistical measures are reduced to scalar values taking either the mean or maximum of the normalized squared differences between the reference and the query.}
    \label{fig:sm_vs_thresholds}
\end{figure}

We evaluate a simplified (though constrained) configuration of FSL-Net that eliminates the Prediction Network and instead relies solely on thresholding the combined statistical measures, computed as the normalized squared differences between the reference and query. In this threshold-based approach, the statistical measures are reduced to a scalar score by taking either the mean or the maximum of these differences. Figure~\ref{fig:sm_vs_thresholds} shows the mean F-1 scores for the SM-only variant of FSL-Net, where feature shift localization is performed entirely through thresholding. The F-1 scores are averaged across datasets, manipulation types, and fractions of manipulated features. The best performance is achieved using the mean-based method with a threshold of 0.002, yielding an optimal F-1 score of 0.307.

\newpage

\subsection{Qualitative Evaluation of FSL-Net on the MNIST Dataset} 

\begin{figure}[!htpb]
    \centering
    \begin{subfigure}[b]{{0.8\textwidth}}
        \centering
        \includegraphics[width=\textwidth]{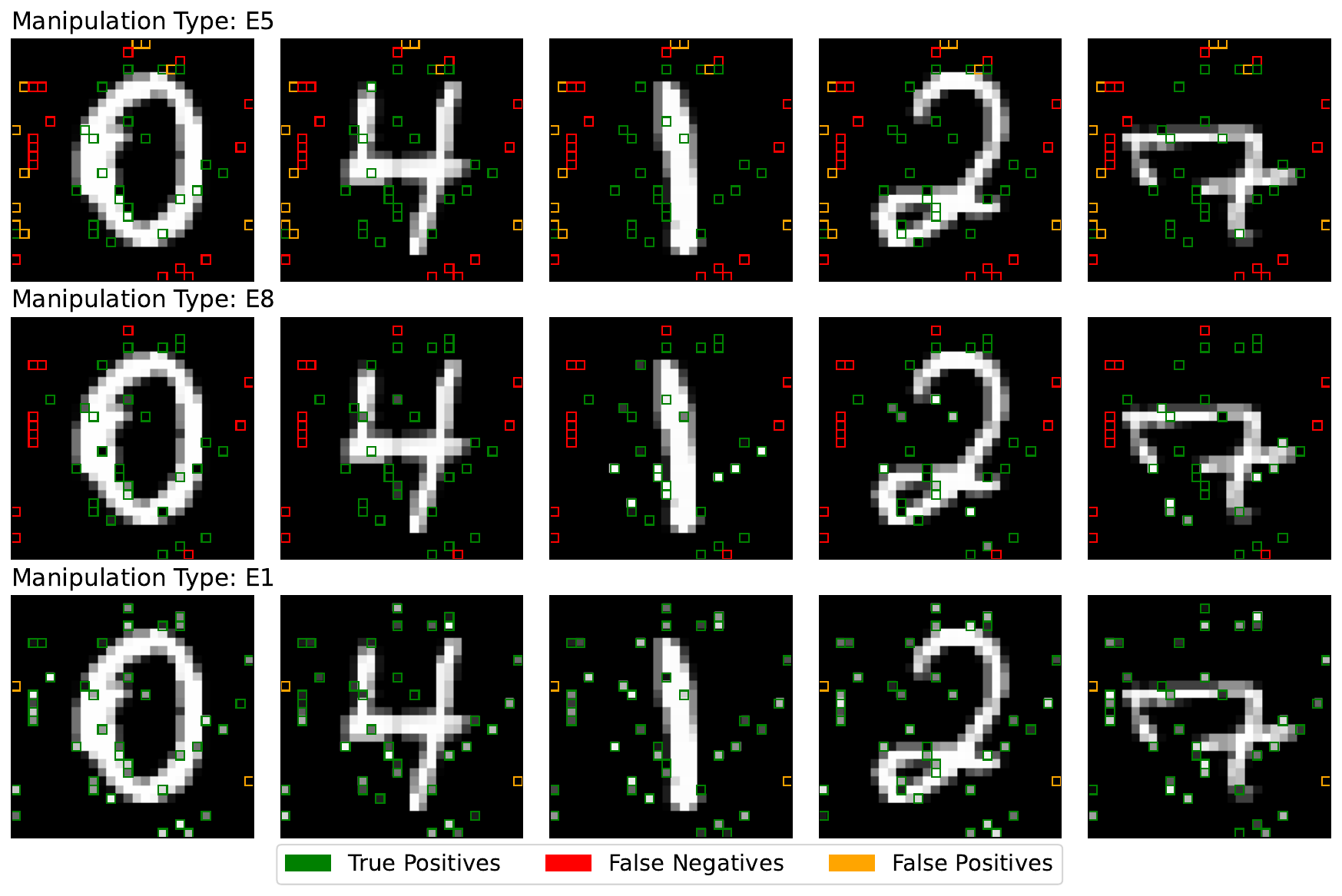}
        \caption{Comparison of FSL-Net output for four examples from the MNIST dataset, with manipulation type E5 (top row), E8 (middle row), and E1 (bottom row). In each case, 5\% of the image features were manipulated. Green rectangles indicate correctly predicted corrupted features, red rectangles denote false negatives, and orange rectangles represent false positives.}
        \label{fig:subfig1}
    \end{subfigure}
    
    \hspace{1em}
    
    \begin{subfigure}[b]{{0.8\textwidth}}
        \centering
        \includegraphics[width=\textwidth]{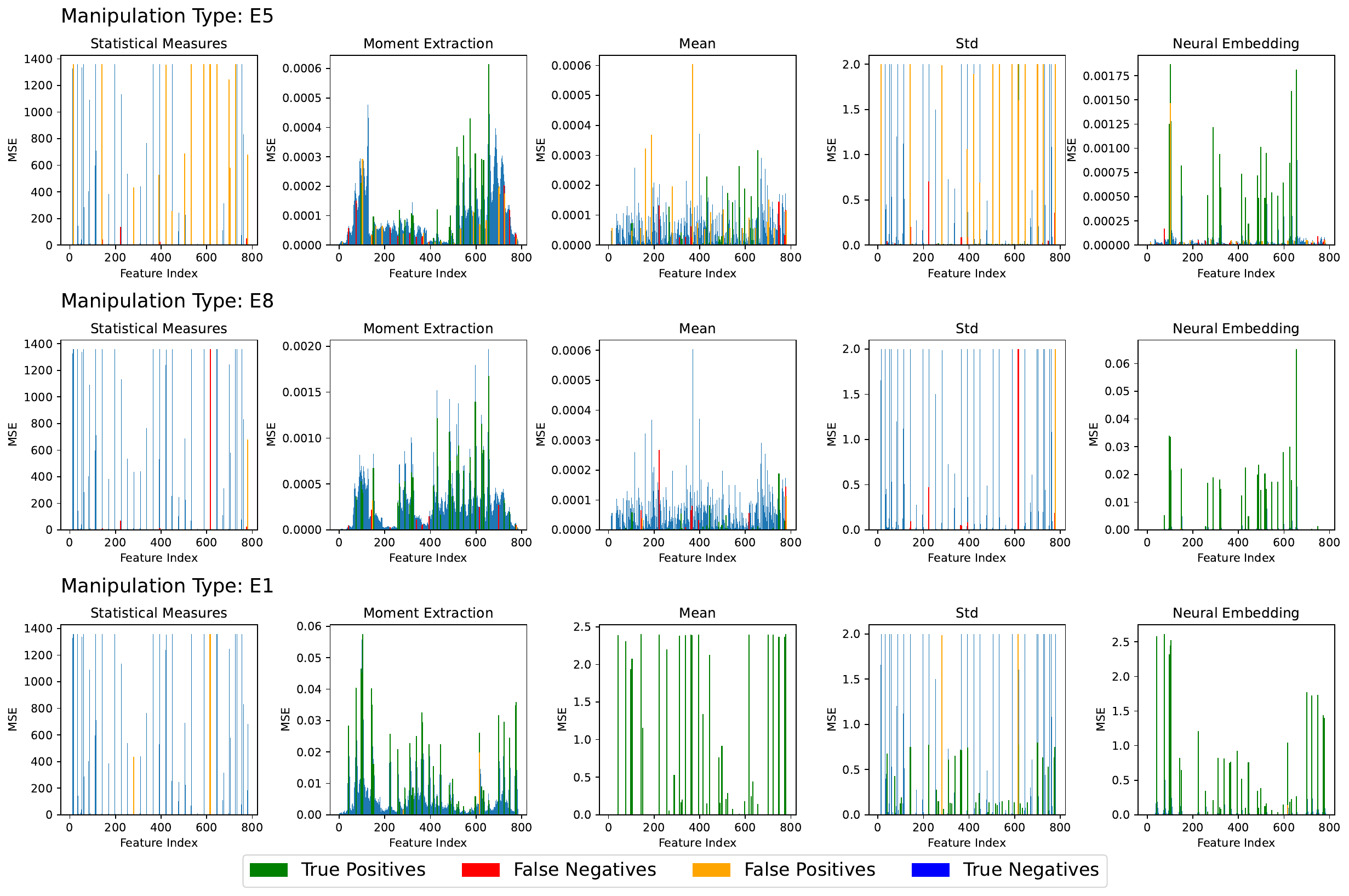}
        \caption{Mean Squared Error (MSE) between reference and query statistical functional maps derived from statistical measures, the Moment Extraction Network, and the Neural Embedding Network. Each bar represents a feature, with colors indicating classification outcomes. The bottom row corresponds to manipulation type E1, while the middle and top rows correspond to the more challenging manipulation types E8 and E5, respectively.}
        \label{fig:subfig2}
    \end{subfigure}
    
    \caption{Qualitative evaluation of FSL-Net on the MNIST dataset.}
    \label{fig:MNIST_qualitative_evalation}
\end{figure}

Figure \ref{fig:MNIST_qualitative_evalation} provides a qualitative assessment of the performance of FSL-Net on the MNIST dataset under various manipulation types. The first subfigure (Figure \ref{fig:subfig1}) illustrates the ability of FSL-Net to identify manipulated features in five sample images. Each row represents a distinct manipulation type applied to modify 5\% of the image features (i.e., pixels), with the top row corresponding to E5, the middle row to E8, and the bottom row to E1. Correctly detected manipulations are highlighted in green, while undetected modifications (false negatives) and misclassified pixels (false positives) are colored red and orange, respectively. The results indicate that FSL-Net effectively detects manipulation E1, while its performance is less consistent for manipulations E5 and E8. The difficulty in detecting E5 comes from its binary rounding of pixel values to 0 or 1, often blending seamlessly into the digit structure. Similarly, E8, which randomly permutes pixel values, disrupts spatial coherence without necessarily creating visually distinct artifacts, making detection more challenging.

The second subfigure (Figure \ref{fig:subfig2}) presents the Mean Squared Error (MSE) between the reference and query statistical functional maps derived from statistical measures (including the mean and standard deviation), the Moment Extraction Network, and the Neural Embedding Network. Each bar represents a distinct feature, with color coding indicating classification outcomes: true positive (green), false negative (red), false positive (orange), and true negative (blue). Manipulation E1 (bottom row) is easily detectable, exhibiting high MSE values, particularly in the statistical functional maps derived from the mean, the Moment Extraction Network, and the Neural Embedding Network. In contrast, more subtle manipulations, such as E5 (top row) and E8 (middle row), display smaller MSE values between reference and query for certain features, making their detection more challenging. Still, several features are correctly classified for these challenging manipulations, with the Neural Embedding Network and the Moment Extraction Network producing the most distinguishable statistical functional maps. The mean and standard deviation measures are ineffective in detecting manipulation E8, as it only affects correlation, with low MSE values for these measures.

\newpage
\end{document}